\documentclass[pdflatex,sn-mathphys-num]{sn-jnl}

\usepackage{graphicx}%
\usepackage{multirow}%
\usepackage{amsmath,amssymb,amsfonts}%
\usepackage{amsthm}%
\usepackage{mathrsfs}%
\usepackage[title]{appendix}%
\usepackage{textcomp}%
\usepackage{manyfoot}%
\usepackage{booktabs}%
\usepackage{algorithm}%
\usepackage{algorithmicx}%
\usepackage{algpseudocode}%
\usepackage{listings}%
\usepackage{adjustbox}
\usepackage{diagbox}
\usepackage{color, soul, colortbl}
\usepackage[dvipsnames]{xcolor}
\usepackage{natbib}
\usepackage{makecell}

\usepackage{pifont}
\usepackage{geometry}
\usepackage{subcaption}
\usepackage{caption}
\usepackage{xurl}
\usepackage{hyperref}
\usepackage{array}
\usepackage{xspace}
\usepackage{colortbl}
\usepackage{tcolorbox}
\usepackage{titlesec}
\usepackage{booktabs}
\usepackage{longtable}
\usepackage{makecell}
\usepackage[htt]{hyphenat}

\usepackage[group-separator={,},group-minimum-digits={3}]{siunitx}

\theoremstyle{thmstyleone}%
\theoremstyle{thmstyletwo}%

\theoremstyle{thmstylethree}%

\newcommand\ours{ConceptCLIP}

\newcommand\ourd{MedConcept-23M}

\begin{document}

\title[Article Title]{An Explainable Biomedical Foundation Model via Large-Scale Concept-Enhanced Vision-Language Pre-training}


\author[1]{\fnm{Yuxiang} \sur{Nie}}
\equalcont{These authors contributed equally to this work.}
\author[1]{\fnm{Sunan} \sur{He}}
\equalcont{These authors contributed equally to this work.}
\author[1]{\fnm{Yequan} \sur{Bie}}
\equalcont{These authors contributed equally to this work.}
\author[1]{\fnm{Yihui} \sur{Wang}}
\author[1]{\fnm{Zhixuan} \sur{Chen}}
\author[1]{\fnm{Shu} \sur{Yang}}
\author[1]{\fnm{Zhiyuan} \sur{Cai}}
\author[1]{\fnm{Hongmei} \sur{Wang}}
\author[1]{\fnm{Xi} \sur{Wang}}
\author[2]{\fnm{Luyang} \sur{Luo}}
\author[3]{\fnm{Mingxiang} \sur{Wu}}

\author[4]{\fnm{Xian} \sur{Wu}}
\author[5, 6]{\fnm{Ronald Cheong Kin} \sur{Chan}}
\author[7]{\fnm{Yuk Ming} \sur{Lau}}

\author[8]{\fnm{Yefeng} \sur{Zheng}}

\author[2]{\fnm{Pranav} \sur{Rajpurkar}}

\author*[1,9,10,11,12]{\fnm{Hao} \sur{Chen}}\email{jhc@cse.ust.hk}

\affil[1]{\orgdiv{Department of Computer Science and Engineering}, \orgname{The Hong Kong University of Science and Technology}, \orgaddress{\city{Hong Kong}, \country{China}}}

\affil[2]{\orgdiv{Department of Biomedical Informatics}, \orgname{Harvard University}, \orgaddress{\city{Boston}, \country{USA}}}

\affil[3]{\orgdiv{Department of Radiology}, \orgname{Shenzhen People’s Hospital}, \orgaddress{\city{Shenzhen}, \country{China}}}

\affil[4]{\orgdiv{Jarvis Research Center}, \orgname{Tencent YouTu Lab}, \orgaddress{\city{Shenzhen}, \country{China}}}

\affil[5]{\orgdiv{Department of Anatomical and Cellular Pathology}, \orgname{The Chinese University of Hong Kong}, \orgaddress{\city{Hong Kong}, \country{China}}}

\affil[6]{\orgdiv{State Key Laboratory of Translational Oncology}, \orgname{The Chinese University of Hong Kong}, \orgaddress{\city{Hong Kong}, \country{China}}}

\affil[7]{\orgdiv{Division of Dermatology, Department of Medicine}, \orgname{Queen Mary Hospital}, \orgaddress{\city{Hong Kong}, \country{China}}}

\affil[8]{\orgdiv{Medical Artificial Intelligence Laboratory}, \orgname{Westlake University}, \orgaddress{\city{Hangzhou}, \country{China}}}

\affil[9]{\orgdiv{Department of Chemical and Biological Engineering}, \orgname{The Hong Kong University of Science and Technology}, \orgaddress{\city{Hong Kong}, \country{China}}}

\affil[10]{\orgdiv{Division of Life Science}, \orgname{The Hong Kong University of Science and Technology}, \orgaddress{\city{Hong Kong}, \country{China}}}

\affil[11]{\orgdiv{State Key Laboratory of Molecular Neuroscience}, \orgname{The Hong Kong University of Science and Technology}, \orgaddress{\city{Hong Kong}, \country{China}}}

\affil[12]{\orgdiv{Shenzhen-Hong Kong Collaborative Innovation Research Institute}, \orgname{The Hong Kong University of Science and Technology}, \orgaddress{\city{Shenzhen}, \country{China}}}

\abstract{
The clinical adoption of artificial intelligence (AI) in medical imaging requires models that are both diagnostically accurate and interpretable to clinicians. While current multimodal biomedical foundation models prioritize performance, their black-box nature hinders explaining the decision-making process in clinically meaningful concepts. Here, we present \textbf{ConceptCLIP}, the first explainable biomedical foundation model that achieves state-of-the-art diagnostic accuracy while delivering human-interpretable explanations across diverse imaging modalities. 
We curate \textbf{MedConcept-23M}, the largest pre-training dataset comprising 23 million image-text-concept triplets across diverse medical modalities, where clinical concepts are derived from the Unified Medical Language System. Leveraging this dataset, we develop ConceptCLIP through a novel dual-alignment approach that simultaneously learns global image-text representations and fine-grained region-concept associations for precise and interpretable medical image analysis. 
We curate the most extensive evaluation benchmark for multimodal biomedical foundation models, covering 52 clinical tasks spanning 10 imaging modalities. Extensive experiments demonstrate that ConceptCLIP outperforms existing state-of-the-art multimodal biomedical foundation models.  
Importantly, ConceptCLIP demonstrates superior diagnostic performance while providing human-understandable explanations validated by clinical experts.
As the first precise and interpretable biomedical foundation model, \ours{} represents a critical milestone toward the widespread clinical adoption of AI,
thereby advancing trustworthy AI in medicine.
}


\maketitle

\section{Introduction}
The clinical adoption of artificial intelligence (AI) systems for medical imaging needs two essential requirements: diagnostic accuracy comparable to specialists and interpretability that supports clinical decision-making~\cite{schwabe2024metric}.
While biomedical foundation models have achieved remarkable accuracy and broad generalizability across diverse imaging modalities~\cite{zhang2023biomedclip,lin2023pmc,eslami2023pubmedclip}, they often function as opaque black boxes, lacking fine-grained, clinically meaningful concepts necessary for explanations in medical practice~\cite{wiens2019no}. 
On the other hand, explainable AI (XAI) methods ~\cite{xai_survey1,xai_survey3,concept_survey,cbm} provide human-interpretable explanations such as medical concepts to help doctors understand the decision-making process of an AI model. However, these methods are constrained by heavy reliance on expert annotations for a specific clinical setting or modality~\cite{daneshjou2022skincon,tsutsui2024wbcatt,derm7pt,luna16}, restricting their broader applicability. The integration of these capabilities—accurate and interpretable medical image analysis with broad generalizability—remains a critical challenge, hindering widespread clinical adoption. 

To address this challenge, we develop ConceptCLIP, the first explainable biomedical foundation model that achieves diagnostic performance superior to state-of-the-art foundation models while providing meaningful explanations across diverse clinical settings.
Our model is pre-trained on \textbf{\ourd{}}, the largest pre-training dataset comprising 23 million image-text-concept triplets derived from 6.2 million scientific articles and enriched with medical terminologies from the Unified Medical Language System (UMLS)~\cite{umls}. As shown in \textbf{Fig. \ref{fig:main} (a)}, \ourd{} includes a diverse range of image types within the field of biomedicine, such as pathology, ultrasound, and X-Ray. Unlike previous datasets with only image-text pairs~\cite{lin2023pmc,zhang2023biomedclip}, \ourd{} is further enhanced with the knowledge of medical concepts (as shown in \textbf{Fig. \ref{fig:main} (b) and (c)}), providing fine-grained textual information paired with various image modalities. 

To effectively capture the relationship between medical images and their corresponding concepts, \ours{} employs a dual alignment approach, consisting of image-text alignment (IT-Align) and region-concept alignment (RC-Align). As shown in \textbf{Fig. \ref{fig:main} (d) and (e)}, IT-Align facilitates the global alignment of medical image and text representations, while RC-Align 
enables local alignment between image regions and their corresponding concepts enhanced by knowledge from UMLS~\cite{umls}. Under this training approach, our model learns to associate medical images with relevant concepts, exploring nuanced clues in the images and concepts that could indicate specific diseases, thereby enhancing the precision of medical image analysis and explainability.

To evaluate the performance of \ours{}, we develop a comprehensive benchmark covering the widest range of tasks, encompassing 52 medical image analysis tasks across 10 image modalities (\textbf{Fig. ~\ref{fig:main} (f)}). Our evaluation includes 36 tasks for medical image diagnosis, incorporating zero-shot, linear probing, and fully fine-tuning settings, to thoroughly assess \ours{} on medical image diagnosis, the fundamental task in medical imaging, across various modalities.  Additionally, we explore the capabilities of \ours{} in other advanced medical tasks, including 15 tasks such as cross-modal retrieval, report generation, visual question answering, and pathology whole-slide image analysis. Furthermore, we conduct extensive explainability analyses across various modalities to evaluate the model's ability for automatic concept annotation, discovering correlations between concepts and diseases, and employing inherently interpretable models.

As depicted in \textbf{Fig. \ref{fig:main} (g), Fig. \ref{fig:cls}, and Fig. \ref{fig:other_tasks}}, \ours{} consistently achieves state-of-the-art results. The outcomes in cross-modal retrieval, zero-shot classification, and visual question answering demonstrate that \ours{} is the leading vision-language pre-training model in the medical domain. Moreover, \ours{} also acts as a well-pretrained vision model in medicine, demonstrating superior performance in linear probing, fully fine-tuning classification, medical report generation, and pathology whole-slide image analysis.
The explainability analyses across various modalities (\textbf{Fig. \ref{fig:main}(g) and Fig. \ref{fig:xai_combined}}) confirm the superior performance of \ours{}.
As the first explainable foundation model in medicine, \ours{} represents a significant milestone toward clinically trustworthy AI, with broad medical applications (\textbf{Fig. \ref{fig:main} (h)}). This advancement paves the way for the practical deployment of reliable, interpretable AI systems in healthcare settings.

In summary, our contributions are fourfold:
\begin{itemize} 
\item We curate \ourd{}, the largest pre-training dataset comprising 23 million medical image-text-concept triplets. This dataset is enriched with standardized medical terminologies from the UMLS, making it a fine-grained and knowledge-enhanced resource for developing trustworthy AI in medicine. 
\item We develop \ours{}, the first explainable biomedical foundation model via concept-enhanced pre-training. It is capable of achieving state-of-the-art performance while offering an interpretable understanding of medical images across diverse modalities. 
\item We conduct an extensive evaluation benchmark of the broadest range across multi-modal biomedical foundation models, including 52 medical image analysis tasks, across 10 image modalities. The results demonstrate that \ours{} achieves state-of-the-art performance among biomedical foundation models, establishing it as the leading medical vision-language pre-training model with strong capability in medical image analysis.

\item We perform comprehensive explainability analysis, exploring how \ours{} enables encoded concept knowledge to provide concept-based explanations, from inherently interpretable model to concept-disease correlation discovery. The results show that ConceptCLIP exhibits promising interpretability, highlighting the model's strong capability to enhance trustworthy AI in the field of medicine.
\end{itemize}

\begin{figure*}[t] 
\centering 
\includegraphics[width=\linewidth,page=1]{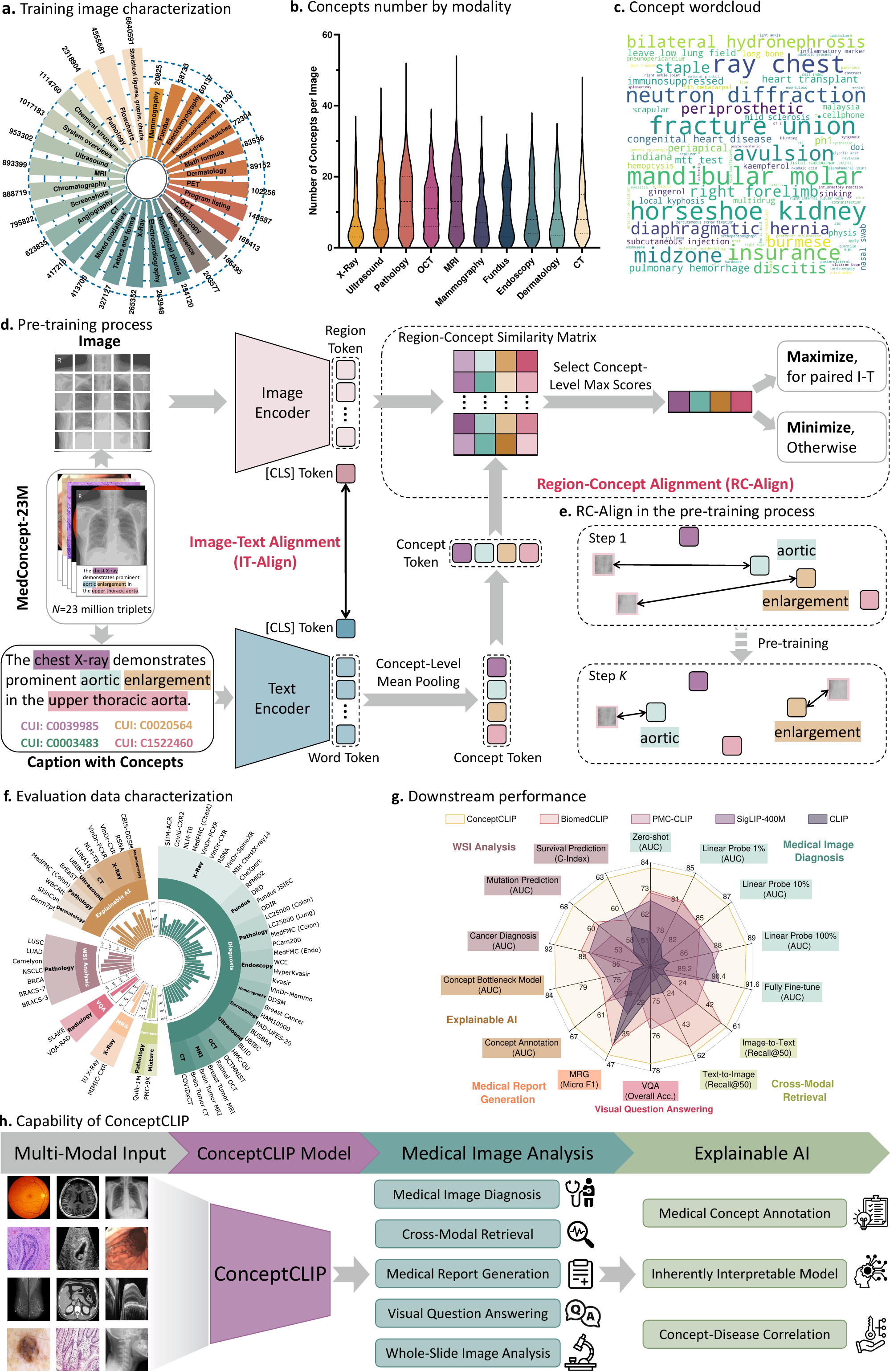} 
\end{figure*}
\clearpage

\begin{figure*}[t]
\caption{ 
\textbf{Overview of the study.} 
\textbf{a,} Estimated distribution of images in \ourd{} dataset by image types. 
\textbf{b,} Distribution of concepts in each medical image modality studied in this work. 
\textbf{c,} The wordcloud of concepts in the \ourd{} dataset. Word size is proportional to its occurrence in \ourd{}. Common nouns and verbs are ignored. Wordclouds for each modality are shown in \textbf{Extended Data Fig.}~\ref{fig:wordcloud_by_modality}.
\textbf{d,} The pre-training process of \ours{} with \ourd{} via an image-text alignment (IT-Align) and region-concept alignment (RC-Align) learning. ``CUI'' refers to ``Concept Unique Identifier'' in UMLS. ``I-T'' denotes ``image and text''.
\textbf{e,} Graphical illustration of the IT-Align in the pre-training process. Clinically related region-concept pairs are gradually learned to align together.
\textbf{f,} Statistical information of the evaluation data across different datasets and modalities. 
\textbf{g,} Comparative performance of \ours{} and other multi-modal biomedical foundation models on various tasks.
\textbf{h,} \ours{} can process diverse modalities and perform versatile tasks in medical image analysis and explainable AI, showcasing its outstanding capability in trustworthy AI.
} \label{fig:main} 
\end{figure*}

\section*{Results}

\subsection*{Curating a large-scale medical image-text-concept dataset}
The \ourd{} dataset is designed to facilitate the pre-training of \ours{}, offering the largest-scale resource of 23 million medical image-text-concept triplets across diverse modalities (\textbf{Fig.} \ref{fig:main} \textbf{(a)-(c)} and \textbf{Extended Data Fig.} \ref{fig:wordcloud_by_modality}), each augmented with UMLS terminology information (as described in the Method section). Specifically, derived from the extensive PubMed Central Open Access Subset (PMC-OA)~\cite{pmc2003}, \ourd{} provides valuable medical knowledge across various domains. By leveraging advanced concept extraction techniques, each image-text pair is enriched with relevant UMLS concepts, resulting in a dataset that is both voluminous and semantically rich. This comprehensive dataset forms the backbone for pre-training \ours{}, enabling it to capture intricate medical relationships and semantics via a dual-alignment pre-training approach, with a global image-text alignment and a region-concept alignment for pre-training (\textbf{Fig. \ref{fig:main} (d) and (e)}).

\subsection*{Comprehensive evaluation benchmark}
Following the pre-training on \ourd{}, \ours{} is subjected to a rigorous evaluation across 52 distinct medical image analysis tasks, as detailed in \textbf{Fig. \ref{fig:main} \textbf{(f)} and Extended Data Table \ref{tab:datasets}}. This is the most extensive benchmark for multi-modal medical foundation models, encompassing 10 different image modalities and a wide range of tasks, including medical image diagnosis, cross-modal retrieval, report generation, visual question answering, and pathology whole-slide image analysis. Importantly, there is no overlap between \ourd{} and the datasets used for evaluation, ensuring an unbiased assessment of \ours{}'s capabilities.
Our evaluation also measures the performance of \ours{} on explainable analyses conducted across various modalities (\textbf{Fig. \ref{fig:main} (d) and (h)}). We highlight \ours{}'s ability to provide plausible explanations, reinforcing its utility in developing trustworthy AI systems in medicine.

\subsection*{Superior diagnostic capabilities of \ours{} across image modalities}

In this section, we present a comprehensive analysis of the performance of \ours{} in medical image diagnosis across various imaging modalities in zero-shot, linear probing, and fully fine-tuning settings (\textbf{Fig.} \ref{fig:cls}). The results demonstrate the superior diagnostic performance of \ours{}, highlighting its outstanding capability in medical image analysis.

\textbf{\ours{} can perform superior zero-shot diagnosis without task-specific training.}
\label{sec:medical_image_diagnosis}
In this experiment, we conduct a systematic evaluation of \ours{}'s ability in the zero-shot setting, where our model is tested on classifying novel classes without requiring additional fine-tuning (\textbf{Fig. \ref{fig:cls} (a), (b) and Extended Data Table \ref{tab:combined_zero_shot_cls_scores}}).
We evaluate \ours{} on 36 datasets spanning 10 medical imaging modalities in zero-shot settings. 
In this experiment, \ours{} is compared with other multimodal foundation models in the general domain and medical domain, which can support diverse modalities. 
By evaluating via AUC (area under the curve) score, our experiment shows that \ours{} consistently outperforms other foundation models across all 10 modalities.
For instance, compared to the previous best-performing model trained on diverse medical data, \ours{} achieves absolute AUC gains of 1.46–10.73\% ($P<0.001$) across modalities (with an average gain of 6.81\%, $P<0.001$), with notable improvements in CT (+10.73\%, $P<0.001$), ultrasound (+10.06\%, $P<0.001$), and fundus (+8.98\%, $P<0.001$). 
Ablation studies (\textbf{Extended Data Table \ref{tab:zero_shot_cls_ablation_scores}}) reveal two critical drivers: concept-enhanced pre-training (contributing 3.76\% of performance gains by incorporating RC-Align in pre-training, $P<0.001$) and top-$K$ local region information for each concept during inference (1.78\%, $P<0.001$). This unified framework, combining diverse modalities with concept-level fine-grained medical knowledge, establishes a foundation for generalizable medical artificial intelligence without modality-specific constraints.

\textbf{\ours{} delivers robust and generalizable medical image representations in linear probing.}
To further investigate the capabilities of the \ours{} image encoder, we evaluate its performance across 30 datasets in medical image analysis. The efficacy of image representation is assessed using linear probing with 1\%, 10\%, and 100\% of the training data, as depicted in \textbf{Fig. \ref{fig:cls} (a), (c)} and detailed in \textbf{Extended Data Tables \ref{tab:combined_linear_probes_1_cls_scores}, \ref{tab:combined_linear_probes_10_cls_scores}, and \ref{tab:combined_linear_probes_100_cls_scores}}. \ours{} consistently surpasses other multi-modal foundation models, which accommodate diverse input modalities, across most modalities and data proportions. For example, with 1\% training data, \ours{} achieves an average AUC of 71.48\% (95\% CI: 62.15\% - 80.44\%) in the X-Ray category, outperforming the second-best model, SigLIP-400M, which scores 67.73\% (95\% CI: 59.20\% - 76.15\%), by 3.75\%  ($p<0.001$). 
This trend persists with 10\% and 100\% training data, where \ours{} demonstrates superior performance among multi-modal foundation models, highlighting its ability to produce 
\begin{figure*}[t] 
\centering 
\includegraphics[width=\linewidth]{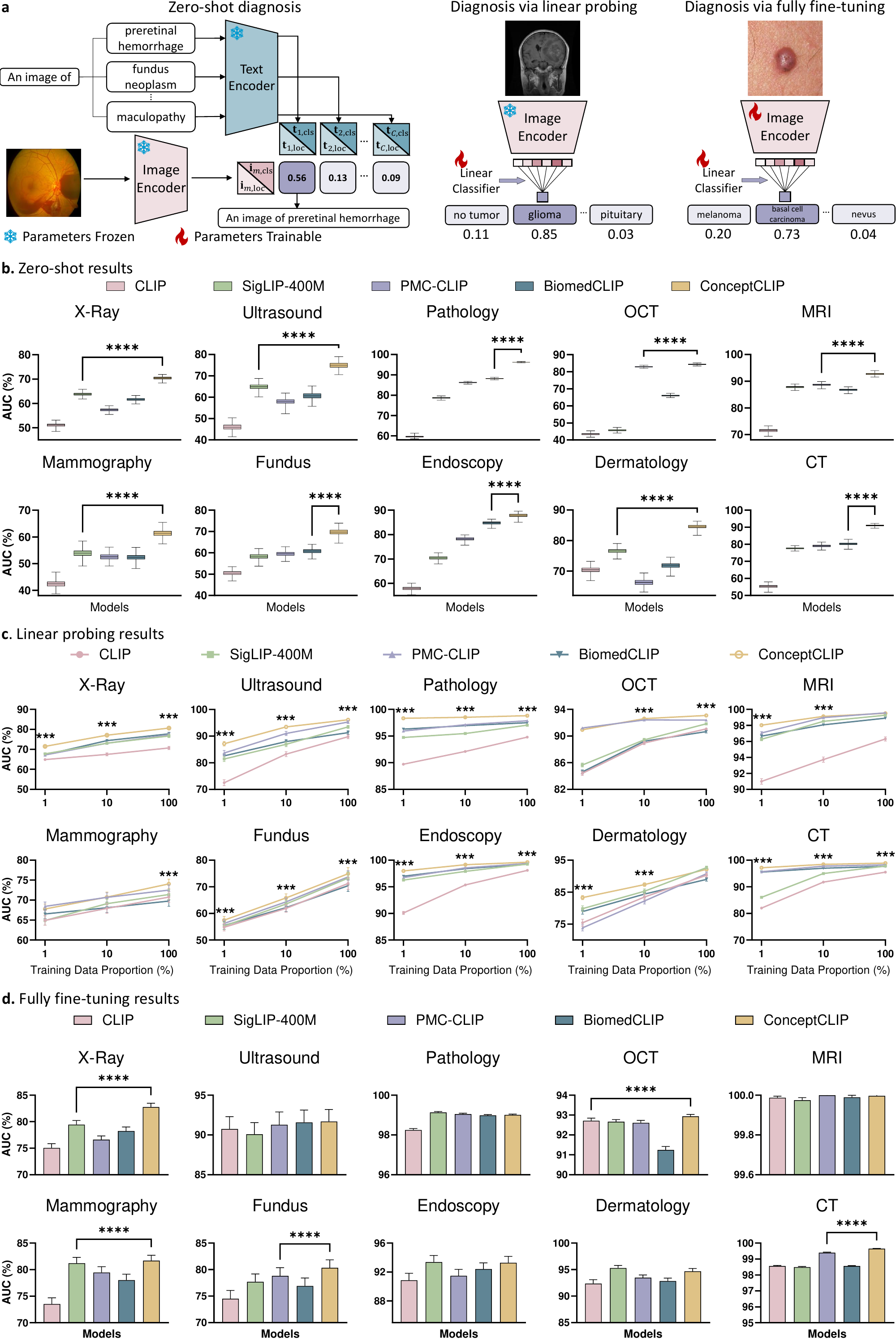} 
\end{figure*}
\clearpage

\begin{figure*}[t]
\caption{ 
\textbf{Medical image diagnosis results.} 
\textbf{a,} The graphical illustration depicts the process of zero-shot diagnosis, diagnosis via linear probing, and diagnosis via fully fine-tuning. For zero-shot diagnosis, $\mathbf{t}_{n, \text{cls}}$ and $\mathbf{t}_{n, \text{loc}}$ denote the global text-level and local concept-level representations of class $n$, respectively, while $\mathbf{i}_{m, \text{cls}}$ and $\mathbf{i}_{m, \text{loc}}$ represent the global and local region-level representations of the $m$-th input image (as described in the Method section).
\textbf{b,} Zero-shot performance is evaluated using 36 datasets across 10 modalities. 
Average AUC scores (\%) are presented for each specific image modality in the plots. Significance levels at which \ours{} outperforms the best competing method are indicated using a two-sided paired \textit{t}-test: ***$P<0.001$; **$P < 0.01$; *$P<0.5$.
\textbf{c,} Linear probing performance is evaluated using 30 datasets across 10 modalities, with the mean ($\pm$s.d.) reported. Experiments considered 1\%, 10\%, and 100\% proportions of the corresponding training set.
\textbf{d,} Fully fine-tuning performance is evaluated using 36 datasets across 10 modalities, with the mean ($\pm$s.d.) reported.} \label{fig:cls} 
\end{figure*} 
\setlength{\parindent}{0pt}high-quality image representations for downstream diagnostic tasks across various modalities. 

\setlength{\parindent}{15pt}
\textbf{\ours{} improves diagnosis performance in fully fine-tuning.}
Fully fine-tuning is a fundamental approach for developing AI models in medical image diagnosis. In this setting, as illustrated in \textbf{Fig. \ref{fig:cls} (a)}, models are fine-tuned on the training set of the specific downstream task prior to evaluation. \textbf{Fig. \ref{fig:cls} (d) and Extended Data Table \ref{tab:combined_fully_finetune_cls_scores}} present the results, where \ours{} consistently demonstrates outstanding performance across all imaging modalities. The model achieves the best performance in 7 out of 10 modalities. 
In the 3 remaining modalities (endoscopy, dermatology, and pathology), \ours{} exhibits marginally lower performance than SigLIP-400M, with AUC differences of 0.08\%, 0.59\%, and 0.12\%, respectively. These minor gaps may arise because SigLIP-400M’s generic pretraining captures certain low-level visual features more effectively in these domains. However, \ours{} enables superior generalization in most diagnostic settings, as evidenced by its 3.29\% AUC gain over SigLIP-400M in X-Ray diagnosis ($P<0.001$).
Overall, these findings suggest that in fully fine-tuning, \ours{} significantly outperforms other multi-modal foundation models, indicating its advantages in this context. 

\subsection*{\ours{} excels in other versatile medical tasks}
This section explores the capabilities of \ours{} beyond medical image diagnosis, examining its performance in cross-modal retrieval, visual question answering, medical report generation, and pathology whole-slide image analysis. The results highlight \ours{}'s versatility and effectiveness across a range of tasks, underscoring its potential to enhance various aspects of medical data analysis.
\begin{figure*}[t] 
\centering 
\includegraphics[width=\linewidth]{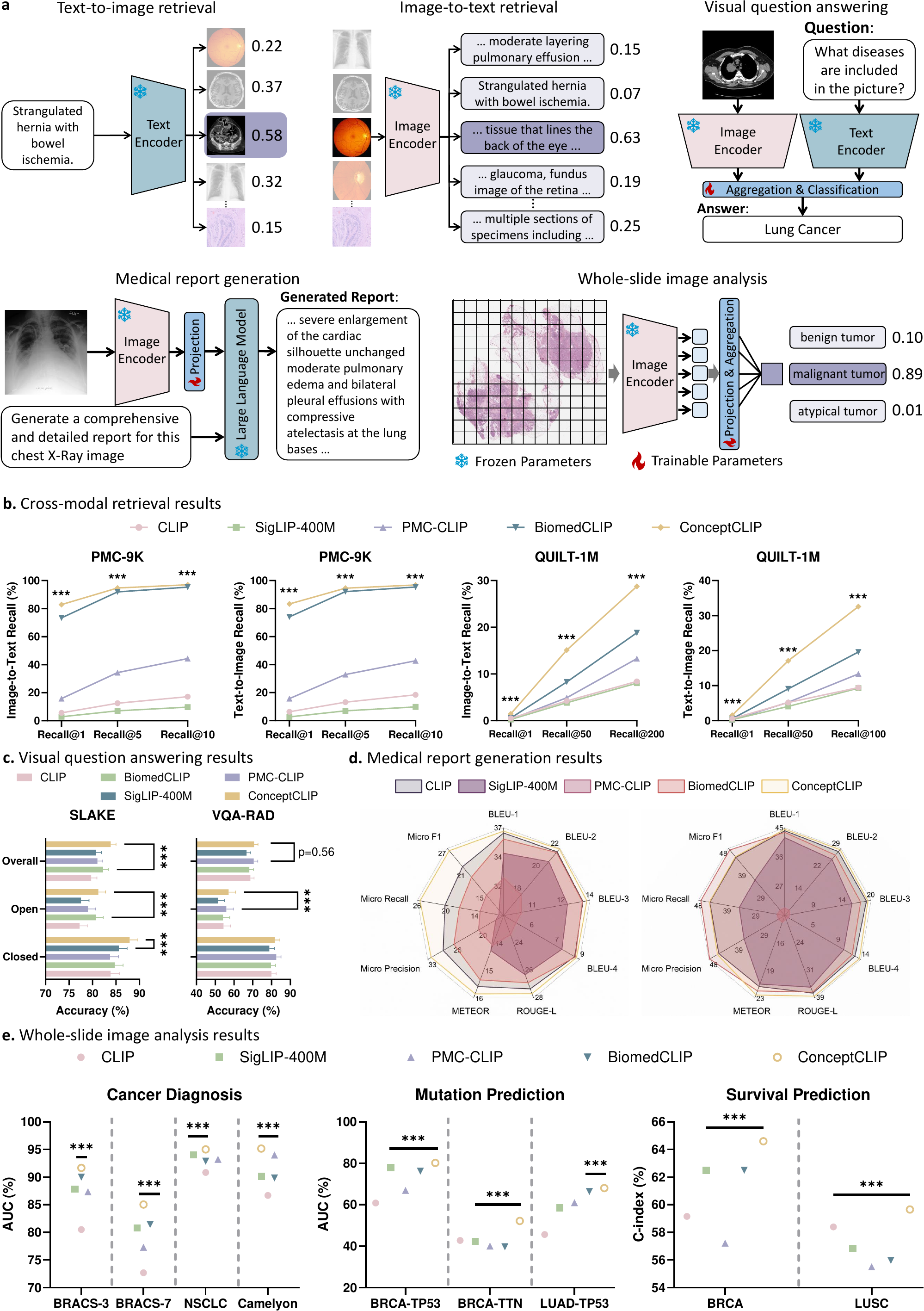} 
\end{figure*}
\clearpage

\begin{figure*}[t]
\caption{ 
\textbf{Results of other medical image analysis tasks.} 
\textbf{a,} This graphical illustration outlines the processes involved in five distinct medical image analysis tasks: text-to-image retrieval, image-to-text retrieval, visual question answering, pathology whole-slide image analysis, and medical report generation. 
\textbf{b,} Performance in cross-modal retrieval is assessed on two datasets, with Image-to-Text Recall (\%) and Text-to-Image Recall (\%) reported for each dataset. 
\textbf{c,} visual question answering tasks are evaluated using accuracy (\%) across two datasets, with the mean ($\pm$s.d.) reported.
\textbf{d,} Performance in medical report generation is presented for two datasets using metrics such as BLEU-1,2,3,4, ROUGE-L, METEOR, Micro Precision, Micro Recall, and Micro F1. 
\textbf{e,} Pathology whole-slide image analysis results are provided across nine tasks within three categories. AUC scores (\%) are reported for cancer diagnosis and mutation prediction, while C-index (\%) is reported for survival prediction, with the mean ($\pm$s.d.) reported.} \label{fig:other_tasks} 
\end{figure*} 

\textbf{\ours{} showcases strong ability in cross-modal image-text retrieval.}
\ours{} demonstrates proficiency in image-to-text and text-to-image retrieval, as shown in \textbf{Fig. \ref{fig:other_tasks} (a)}. The text encoder of \ours{} generates a representation of the input text, which is used to calculate similarities with representations of candidate images encoded by the image encoder. We evaluate this capability using the PMC-9K (we curated it as described in the Method section) and QUILT-1M~\cite{ikezogwo2024quilt} datasets, comparing \ours{} with other multi-modal foundation models, as detailed in \textbf{Fig. \ref{fig:other_tasks} (b), Extended Data Table \ref{tab:retrieval_metrics_pmc} and \ref{tab:retrieval_metrics_quilt}}. \ours{} achieves the highest Recall scores in both Image-to-Text and Text-to-Image tasks. Specifically, in the PMC-9K dataset, \ours{} attains a Recall@1 of 82.85\% (95\% CI: 82.05\% - 83.60\%) for Image-to-Text retrieval, significantly outperforming BiomedCLIP, which scores 73.41\% (95\% CI: 72.53\% - 74.32\%), by 9.44\% ($P<0.001$). 
In QUILT-1M, \ours{} achieves a Recall@200 score of 32.50\% (95\% CI: 31.79\% - 33.45\%) for Text-to-Image retrieval, representing a 12.88\% increase ($P<0.001$) over BiomedCLIP, which scores 19.62\% (95\% CI: 18.91\% - 20.31\%). This superior performance demonstrates \ours{}'s proficiency in understanding and linking visual and textual information, a crucial capability for developing integrated medical information retrieval systems.

\textbf{\ours{} is proficient in medical visual question answering.}
The visual question answering (VQA) task is crucial for evaluating a model's ability to comprehend and respond to complex medical queries. As depicted in \textbf{Fig. \ref{fig:other_tasks} (a)}, a medical image and a question are encoded by the image and text encoders of a foundation model, respectively. These are then processed by a fusion module to calculate the similarity to candidate answers, ultimately selecting the most appropriate answer (described in the Method section). We analyzed the VQA performance of \ours{} using the SLAKE and VQA-RAD datasets. As shown in \textbf{Fig. \ref{fig:other_tasks} (c) and Extended Data Table \ref{tab:vqa_scores}}, \ours{} achieves the highest overall accuracy in both datasets, with 83.86\% (95\% CI: 81.53\% - 85.96\%) on SLAKE and 70.70\% (95\% CI: 66.73\% - 74.73\%) on VQA-RAD, outperforming other multi-modal foundation models. These results indicate that \ours{} effectively comprehends and responds to complex medical queries, demonstrating its potential as a decision-support tool in clinical environments where accurate medical decision-making is essential.

\textbf{\ours{} improves the quality of medical report generation.}
Medical report generation is a critical task for summarizing complex medical information. As illustrated in \textbf{Fig. \ref{fig:other_tasks} (a)}, the pre-trained image encoder extracts image features, which are concatenated with text embeddings and processed by a large language model to generate the final report (the Method section). \textbf{Fig. \ref{fig:other_tasks} (d) and Extended Data Table \ref{tab:mrg_scores}} present the performance of \ours{} on the medical report generation task using the MIMIC-CXR~\cite{johnson2019mimic} and IU X-Ray~\cite{demner2016preparing} datasets. \ours{} consistently outperforms or achieves competitive performance with other models on most metrics for these datasets. Notably, in MIMIC-CXR, \ours{} achieves a Macro F1 score of 26.38\% (95\% CI: 25.37\% - 27.41\%), surpassing the second-best result by 4.28\% 
($P<0.001$), indicating its superior ability to generate accurate and coherent medical reports. This performance suggests that \ours{} effectively understands and summarizes complex medical information, providing valuable support in clinical documentation. However, in IU-X-Ray, the Micro Precision, Recall, and F1 scores of \ours{} were not as high as BiomedCLIP. This discrepancy is attributed to the imbalance in the IU X-Ray dataset, where most cases are normal. A model that consistently outputs normal results can achieve high Micro Precision, Recall, and F1 scores, indicating a bias in the dataset that does not fully reflect a model's performance.

\textbf{\ours{} enables precise pathology whole-slide image analysis.}
Whole-slide image analysis is a vital task for enhancing diagnostic accuracy in histopathology. As depicted in \textbf{Fig. \ref{fig:other_tasks} (a)}, a whole-slide image is divided into patches, each of which is processed by the pre-trained vision encoder. The extracted visual features are aggregated for downstream classification tasks. As shown in \textbf{Fig. \ref{fig:other_tasks} (e) and Extended Data Table \ref{tab:wsi_cls_scores}}, \ours{} demonstrates exceptional performance across various tasks, including cancer diagnosis, mutation prediction, and survival prediction. For instance, in the cancer diagnosis task on the BRACS-3 dataset, \ours{} achieves an AUC score of 91.65\% (95\% CI: 87.18\% - 96.12\%), outperforming all other models 
($P<0.001$).
Similarly, in the survival prediction task for LUSC, \ours{} records the highest C-index of 59.66\% (95\% CI: 50.31\% - 69.01\%). These results highlight \ours{}'s capability to process and analyze pathology images of a humongous size, making it a promising tool for enhancing diagnostic accuracy in histopathology.

\subsection*{\ours{} excels in advancing explainable AI}

In this section, we assess the explainability of ConceptCLIP. As the first explainable biomedical foundation model, ConceptCLIP delivers robust and generalizable explainable AI (XAI) performance, consistently providing clinically meaningful concept-level explanations across diverse medical imaging modalities and diagnostic scenarios. Specifically, we analyze the model explainability from three key perspectives: medical concept annotation, inherently interpretable model employment, and concept-disease correlation discovery.

\begin{figure*}[t] 
\centering 
\includegraphics[width=\linewidth,page=1]{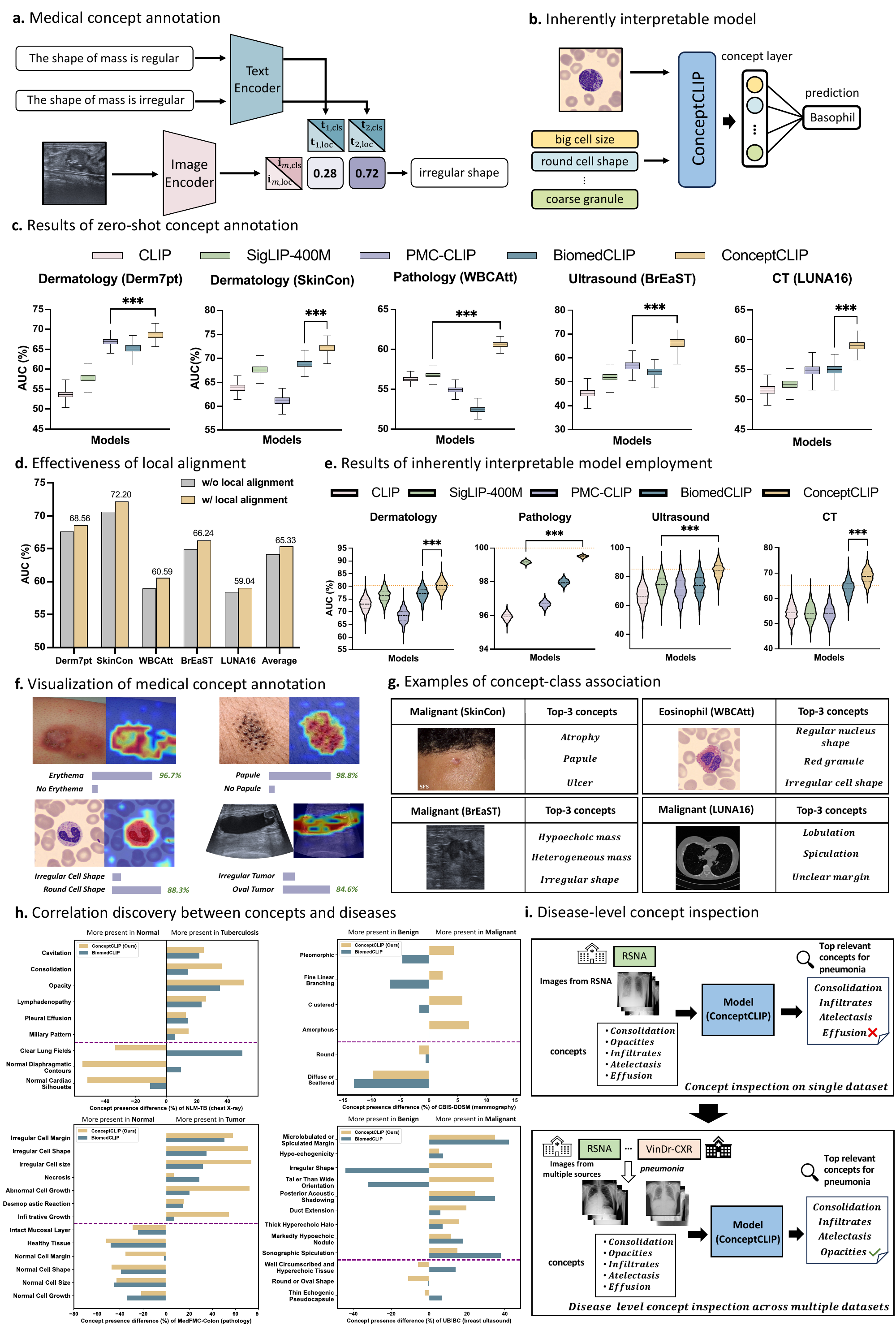} 
\end{figure*}
\clearpage

\begin{figure*}[t]
\caption{\textbf{Explainability experiment results.} \textbf{a}, The illustration of zero-shot medical concept annotation. \textbf{b}, The illustration of an inherently interpretable model built upon ConceptCLIP, where a linear concept layer is adopted before the final prediction. \textbf{c}, The comparison results of zero-shot concept annotation on datasets of four modalities, which have fine-grained concept labels, with 95\% CI reported. 
\textbf{d}, The effectiveness of the local alignment strategy, where yellow and gray bars represent the medical concept annotation results with local alignment and without local alignment during inference, respectively. \textbf{e}, Performance comparison of inherently interpretable models built upon medical vision-language models on disease diagnosis tasks, with 95\% CI reported. In addition, the orange horizontal line indicated in each sub-figure represents the result of fully fine-tuned black-box models. \textbf{f}, Visualization of zero-shot medical concept annotation. For each example, the focused regions are highlighted based on the gradient of image-concept similarities, where the predicted probabilities of concepts are also presented. \textbf{g}, Examples of concept-class association offered by the inherently interpretable model built upon ConceptCLIP. \textbf{h}, Results of discovery for correlations between concepts and diseases on datasets of various modalities. According to the clinical fact, concepts located above the purple horizontal line have stronger associations with the category indicated on the upper right of each figure (e.g., Tuberculosism and Malignant), while concepts below the purple line show greater relevance to the left category (e.g., Normal and Benign). We calculate the concept presence difference to obtain which concepts are more likely to appear in which image set using the zero-shot concept annotation capability of our model. \textbf{i}, Disease-level concept inspection. For a specific disease, disease-level concept inspection conducts global concept-disease association analysis for images from multiple datasets to get more accurate and consistent concept-based explanations.}
\label{fig:xai_combined}
\end{figure*}

\textbf{\ours{} showcases strong ability in medical concept annotation.}
\label{sec:explainable_ai}
Concepts are high-level and human-interpretable attributes or symptoms within images, which are essential for doctors or medical experts to diagnose diseases. How to effectively predict concepts for medical images of different modalities is an important and challenging research direction for XAI. In this experiment, 
we evaluate the performance of zero-shot medical concept annotation in various modalities, including dermatology, pathology, breast ultrasound, and chest CTs. The illustration of medical concept annotation of our model is shown in \textbf{Fig. \ref{fig:xai_combined} (a)}, where the concept presence outcome is determined by identifying the candidate prompt of concepts with the highest similarity score to the input image. Here, $\mathbf{t}_{c,\text{cls}}$ and $\mathbf{t}_{c,\text{loc}}$ respectively denote the global and local text representations for each candidate prompt of concepts, while $\mathbf{i}_{m,\text{cls}}$ and $\mathbf{i}_{m,\text{loc}}$ signify the global and local representation of the input image, respectively. The results are reported in \textbf{Fig. \ref{fig:xai_combined} (c) and Extended Data Table \ref{tab:zs_concept_anno}}. Specifically, our model achieves state-of-the-art performance, outperforming other compared multimodal foundation models. The average performance of ConceptCLIP achieves 10.9\% and 10.4\% relative performance boost ($P<0.001$) over the results of PMC-CLIP and BiomedCLIP, respectively, 
highlighting its superior capability in medical concept annotation. To demonstrate the effectiveness of local alignment for concept annotation, we conduct ablation experiments and report the results in \textbf{Fig. \ref{fig:xai_combined} (d)}, where our method with local alignment between local image and concept features during inference consistently outperforms baselines without local alignment. In \textbf{Fig. \ref{fig:xai_combined} (f)}, we present the image-concept visualization to help better understand the model decision by highlighting the focus region of our model for specific concepts. From the experimental results, it can be observed that our model has the capability to identify clinical concepts across various modalities, thereby holding the potential to facilitate the development of trustworthy models and medical image datasets with fine-grained concept annotations.

\setlength{\parindent}{15pt}\textbf{\ours{} can be employed as an inherently interpretable model.}
Within explainable AI, various approaches have been proposed to explain neural networks. Some methods utilize saliency maps \cite{cam,grad_cam,x_mir} to highlight the contribution of each pixel or region in the model’s predictions, while some focus on feature interactions \cite{tsang2017detecting} and influence functions \cite{koh2017understanding} to explain the model. However, these post-hoc XAI techniques may not be faithful enough to truly reflect the model decision-making process \cite{rudin2019stop}. Therefore, building inherently interpretable models such as Concept Bottleneck Models (CBMs) \cite{cbm} gains increasing attention since these models allow doctors and patients to easily understand the model decisions, hence holding the potential to advance trustworthy AI in medicine. Specifically, CBM, as an inherently interpretable model, first predicts the concepts present in the given images and then makes the final prediction based on the concepts through a linear layer, as shown in \textbf{Fig. \ref{fig:xai_combined} (b)}. For the concept bottleneck layer, each neuron represents one pre-defined human-understandable concept. Thus, the weights of the layer can be regarded as the contribution of each concept to the final class (i.e., concept-class association). In this paper, we report the results of inherently interpretable models built upon ConceptCLIP and other multimodal foundation models on disease diagnosis, as shown in \textbf{Fig. \ref{fig:xai_combined} (e) and Extended Data Table \ref{tab:cbm}}. Our proposed model significantly outperforms all other compared foundation models, achieving a 6.13\% improvement ($P<0.001$) in AUC score over the second-best method. Moreover, we compare the results of the employed inherently interpretable models with black-box models, as indicated using the orange line in \textbf{Fig. \ref{fig:xai_combined} (e)}. Specifically, we observe that black-box models consistently outperform other medical foundation models with concept bottlenecks. However, our model performs comparably and even outperforms fully fine-tuned black-box models in most settings. 
For example, ConceptCLIP with CBM gains 5.58\% performance boost compared to the fully fine-tuned black-box models on the LUNA16 dataset. 
Additionally, the results of concept-class association are shown in \textbf{Fig. \ref{fig:xai_combined} (g)}, where we present the top 3 relevant concepts for each selected disease across four image modalities, ranked by their weights in the concept linear layer, paired with an example image of the specific disease from each considered dataset. The results of inherently interpretable model employment demonstrate that ConceptCLIP achieves promising diagnosis performance while offering concept-based explanations, holding the potential to be employed as an inherently interpretable model and be practically used in disease diagnosis with explanations.

\textbf{\ours{} can discover correlations between concepts and diseases.}
One of the most promising applications of XAI in medicine is the discovery and interpretation of correlations between medical concepts and diseases. By understanding the relationships between various medical concepts (e.g., symptoms and observations in medical images) and diseases, the AI model can better assist doctors in diagnosis with more reliable and accurate concept-based explanations. In addition, given the reference of which concepts are more relevant to which diseases, medical experts and patients can perform possible disease screening more easily, leading to earlier and more precise diagnoses and treatments.
Previous research \cite{monet} conducts data auditing to validate the correlations between concepts and diseases within data. However, it can only be applied to analyze skin images, which is insufficient for trustworthy AI in general medicine. In contrast, ConceptCLIP can discover the correlations and analyze medical images of various modalities by measuring the concept presence difference of different image sets (as described in the Method section) for medical datasets without fine-grained concept labels. Specifically, to demonstrate the concept-disease correlation discovery ability of our model, we conduct comparison experiments on datasets including modalities of chest X-rays, mammography, pathology, and ultrasound. The results are shown in \textbf{Fig. \ref{fig:xai_combined} (h)}. We observe that ConceptCLIP successfully identifies differentially present concepts in different datasets. For example, in the experiment of the NLM-TB dataset \cite{jaeger2014two}, our model concludes that the concepts \textit{cavitation, consolidation, opacity, lymphadenopathy}, etc., are more likely to present in the image set of tuberculosis while \textit{clear lung fields, normal diaphragmatic contours, }etc., present more in normal chest X-rays, which are consistent with the clinical findings.

In addition to the analysis for a single dataset, a global concept inspection for a disease beyond the granularity of the dataset level is also essential for human users to better comprehend how the model deciphers the disease in terms of human-interpretable concepts. The inspection results may be biased when only one dataset is considered, due to potential spurious correlations within the data and the limited number of data samples. The disease-level concept inspection result of our model is shown in \textbf{Fig. \ref{fig:xai_combined} (i)}, with pneumonia as the selected disease. It can be observed that when only one dataset (e.g., RSNA from the National Institutes of Health) is inspected, the output top relevant concepts for pneumonia include \textit{consolidation, infiltrates, atelectasis,} and \textit{effusion}. The result changes to the concepts of \textit{consolidation, infiltrates, atelectasis}, and \textit{opacities} when more images of pneumonia extracted from multiple datasets are considered (e.g., VinDr-CXR from the Hospital 108 and the Hanoi Medical University Hospital, VinDr-PCXR from Phu Tho Obstetric \& Pediatric Hospital). The results of disease-level concept inspection across multiple datasets are more consistent with the ground-truth concepts given by certified radiologists. The observations demonstrate the potential of our model to inspect specific diseases with more accurate and consistent global concept-based explanations across various datasets.

\section*{Discussion}\label{sec:discussion}
In this work, we introduce \ours{}, the first explainable biomedical foundation model that excels in both medical image analysis and explainability, marking a significant advancement in trustworthy AI for healthcare. Leveraging \ourd{}, the largest pre-training dataset with 23 million image-text-concept triplets, \ours{} is designed to understand and interpret complex medical data across diverse modalities. This extensive dataset, derived from the PubMed Central Open Access Subset and enriched with UMLS concepts, provides a rich foundation that allows \ours{} to capture intricate relationships and semantics within medical data. This capability is critical for tasks such as medical image analysis and concept-driven explainable AI, where understanding nuanced details in medical images is essential.

\textbf{Our dual alignment strategy enhances precision and interpretability.} The innovative dual alignment approach in \ours{}—comprising image-text alignment (IT-Align) and region-concept alignment (RC-Align)—is instrumental in enhancing the model's performance. IT-Align integrates global image and text representations, ensuring that the model can comprehend the broader context of medical images. Meanwhile, RC-Align employs UMLS concept knowledge to reinforce local region-concept associations, allowing for more precise and nuanced understanding of specific image areas. This dual strategy significantly improves \ours{}'s ability to accurately associate medical images with relevant concepts, thereby enhancing both analytical precision and interpretability. Experimental results demonstrate that this approach allows \ours{} to outperform other models in zero-shot settings across various modalities, highlighting its robust capability to generalize without additional training.

\textbf{\ours{} outperforms existing multi-modal medical foundation models across a broad benchmark.} Evaluated on the most comprehensive benchmark with 52 tasks across 10 modalities, \ours{} achieves state-of-the-art results among medical multi-modal foundation models (\textbf{Fig. \ref{fig:main} (d) and (e)}). This extensive evaluation includes diverse tasks such as medical image diagnosis, cross-modal retrieval, and visual question answering, ensuring a rigorous assessment of the model's capabilities. \ours{} consistently surpasses other models in key performance metrics, demonstrating its versatility and robustness. For instance, in zero-shot classification tasks (\textbf{Fig. \ref{fig:cls} (b) and Extended Data Table \ref{tab:combined_zero_shot_cls_scores}}), \ours{} achieves significant AUC gains compared to other models, underscoring its ability to handle a wide range of medical imaging tasks without the need for extensive fine-tuning.

\textbf{Explainability is a core strength of \ours{}, fostering trust in AI.} A standout feature of \ours{} is its ability to provide concept-based explanations alongside diagnostic results. This transparency is critical for clinical adoption, as it enhances trust among healthcare professionals and patients. By elucidating the rationale behind AI predictions, \ours{} supports reliable decision-making. Our experiments show that \ours{} excels in medical concept annotation (\textbf{Fig. \ref{fig:xai_combined} (c) and Extended Data Table \ref{tab:zs_concept_anno}}) and can effectively highlight relevant image regions (\textbf{Fig. \ref{fig:xai_combined} (c)}), providing interpretable insights that are crucial for developing trustworthy AI systems in medicine.

\textbf{The scale of \ours{} is pivotal to its success.} The large scale of \ours{}, with a vision encoder of 400 million parameters—the largest among all vision encoders in multi-modal medical foundation models—and its pre-training on an extensive dataset with 23 million image-text-concept triplets are key to its exceptional performance. This underscores the importance of large models and datasets in advancing medical AI, suggesting that future improvements will likely stem from further scaling these resources, thereby enhancing AI capabilities and trustworthiness. The comprehensive nature of \ourd{} ensures that \ours{} can generalize across a wide array of tasks and modalities, setting a new benchmark for what is achievable in medical AI.

\textbf{Limitations and future directions.} While \ours{} demonstrates significant advancements in medical vision-language integration, several limitations warrant consideration. First, despite the extensive scale of \ourd{}, the dataset may still lack representation of certain rare diseases or conditions, potentially limiting the model's ability to generalize across all medical scenarios. This highlights the need for continuous dataset expansion and diversification to ensure comprehensive coverage of medical knowledge. Second,  while \ours{} excels in providing concept-based explanations, it requires pre-defined concepts for analysis. Its open-set capability, essential for biomarker discovery, remains unexplored and is a promising future direction. Finally, while \ours{} has been evaluated on a broad benchmark, more real-world clinical validation is essential to fully assess its utility and impact in diverse healthcare environments. This necessitates collaboration with clinical practitioners to ensure the model meets practical needs and integrates seamlessly into existing workflows.

\clearpage

\section*{Method}
\label{sec:method}
In this section, we first provide a detailed description of the construction of medical image-text-concept triplets. Subsequently, we explore how these triplets can be utilized for model pre-training.

 \subsection*{\ourd{}: a dataset of 23 million medical image-text-concept triplets}

In this section, we detail the construction of the \ourd{} dataset, which encompasses two primary stages: dataset collection and concept extraction. Following this, we examine the curated dataset statistics.

\subsubsection*{Dataset collection}
To collect large-scale medical image-text pairs for pre-training a robust medical vision-language model, we focus on the PubMed Central Open Access Subset (PMC-OA)~\cite{pmc2003}, which contains 6,246,351 articles under licenses permitting reuse (as of August 19, 2024). The articles are downloaded from \url{https://ftp.ncbi.nlm.nih.gov/pub/pmc/oa_package/}. Using PubMed Parser~\cite{achakulvisut2020pubmed}, we extract 23,289,898 image-text pairs, allocating 23,000,000 for pre-training and 289,898 for evaluation. 

The pre-training set retains all image types, including both clinical images (e.g., radiology) and scientific figures (e.g., diagrams), as the latter provide valuable structural and semantic signals (e.g., anatomical relationships, biomarker trends) beneficial for representation learning~\cite{zhai2023sigmoid}. For evaluation purposes, we establish PMC-9K through a rigorous curation process: first applying automated filtering via an image classifier~\cite{subramanian-2020-medicat} and InternVL~\cite{chen2024internvl}, followed by manual verification ($> 96\%$ relevance, $\kappa=0.89$) to ensure clinical applicability.

\subsubsection*{Concept Extraction}
We develop a rigorous UMLS concept extraction pipeline comprising three stages: (1) Initial biomedical entity recognition using SciSpacy~\cite{neumann-etal-2019-scispacy}, whose UMLS-linked named entity recognizer demonstrates state-of-the-art performance (0.84 F1) on the BC5CDR benchmark~\cite{li2016biocreative}; (2) Semantic disambiguation via SciSpacy's UMLS linker, which computes concept similarity scores using a pretrained clinical BERT model, retaining only matches exceeding the empirically optimized threshold of 0.8 (achieving 95\% precision on internal validation); and (3) Contextual validation with PubMedBERT~\cite{achakulvisut2020pubmed} to ensure concept-caption coherence. This process yields \ourd{}—a dataset of 23 million precisely aligned image-text-concept triplets.

\subsubsection*{Dataset statistics}
As illustrated in \textbf{Fig. \ref{fig:main} (a), (b), and (c)}, the curated dataset encompasses medical images from various domains, including pathology, ultrasound, X-Ray, and other medical image modalities. These images form the foundation for pre-training a robust medical vision-language model capable of handling diverse image modalities. We treat standard X-ray (denoted as ``X-Ray'') and mammography (breast-optimized X-ray, denoted as ``Mammography'') as distinct modalities due to fundamental differences in image resolution, contrast requirements, and anatomical focus—factors critical for diagnostic applications.

\subsection*{\ours{}: concept-enhanced contrastive language-image pre-training in medicine}

In this section, we first introduce the architecture of the image and text encoders in \ours{}, followed by a detailed description of the training objectives, including global image-text alignment (IT-Align) and local region-concept alignment (RC-Align).
We consider a dataset of \(N\) image-text-concept triplets, \(\mathcal{D} = \{(I_1, T_1, G_1), \ldots, (I_N, T_N, G_N)\}\), where \(I_m\) denotes the \(m\)-th medical image, \(T_m\) is the corresponding caption, and \(G_m\) is the set of concepts extracted from \(T_m\). Our goal is to develop a medical vision-language model (VLM) that learns aligned representations of medical images and text through a dual-encoder framework, enabling robust analysis and explainability in healthcare applications.

\subsubsection*{Image and Text Encoding}
\textbf{Image Encoder.} The image encoder \(f(\cdot)\) is a Vision Transformer (ViT) that processes an input image \(I_m\) into a structured representation. Following SigLIP~\cite{zhai2023sigmoid}, the encoder outputs:  
\begin{equation} 
\mathbf{i}_m = f(I_m) = [\mathbf{i}_{m,\text{cls}}; \mathbf{i}_{m,1}; \mathbf{i}_{m,2}; \ldots; \mathbf{i}_{m,r}] \in \mathbb{R}^{(r+1) \times h},
\end{equation}  
where \(\mathbf{i}_{m,\text{cls}} \in \mathbb{R}^h\) is the global [CLS] token embedding, \(\mathbf{i}_{m,1:r} \in \mathbb{R}^{r \times h}\) are patch-level region features, \(r\) is the number of image regions, and \(h\) is the hidden dimension.  

\textbf{Text Encoder.} The text encoder \(g(\cdot)\) is a PubMedBERT model~\cite{achakulvisut2020pubmed} that maps a text \(T_m\) into tokenized embeddings:  
\begin{equation}
\mathbf{t}_m = g(T_m) = [\mathbf{t}_{m,\text{cls}}; \mathbf{t}_{m,1}; \mathbf{t}_{m,2}; \ldots; \mathbf{t}_{m,s}] \in \mathbb{R}^{(s+1) \times h},
\end{equation}  
where \(\mathbf{t}_{m,\text{cls}} \in \mathbb{R}^h\) is the global text [CLS] token, \(\mathbf{t}_{m,1:s} \in \mathbb{R}^{s \times h}\) are token-level embeddings, and \(s\) is the text length.  

The image and text encoders are initialized with pre-trained weights from SigLIP and PubMedBERT, respectively, and fine-tuned during training to align medical semantics across modalities.

\subsubsection*{Training approach}
In this section, we introduce a dual alignment training approach, which consists of two components: image-text alignment learning and region-concept alignment learning (\textbf{Fig. \ref{fig:main} (d) and (e)}).

\textbf{Image-text alignment learning (IT-Align).} We employ a sigmoid loss function following SigLIP~\cite{zhai2023sigmoid}, to perform image-text alignment:
\begin{equation}
    \mathcal{L}_{\text{IT-Align}}=-\frac{1}{|\mathcal{B}|}\sum^{|\mathcal{B}|}_{m=1}\sum^{|\mathcal{B}|}_{n=1}\text{log}\frac{1}{1+e^{{z^{m,n}(-t_g\mathbf{x}_m\cdot\mathbf{y}_n+b_g)}}}.
    \label{eq:it-align}
\end{equation}

Here, $|\mathcal{B}|$ represents the size of the mini-batch of image-text pairs. The parameters $t_g$ and $b_g$ are the logit scale and logit bias for the global image-text alignment, respectively. The variable $z_{m,n}$ equals 1 if the image and text are paired (e.g., $m=n$), and -1 otherwise (i.e., $m\neq n$). The normalized representations of the $m$-th image and the $n$-th text are given by $\mathbf{x}_m=\frac{\mathbf{i}_{m,\text{cls}}}{||\mathbf{i}_{m,\text{cls}}||}$ and $\mathbf{y}_n=\frac{\mathbf{t}_{n,\text{cls}}}{||\mathbf{t}_{n,\text{cls}}||}$, respectively. 

\textbf{Region-concept alignment learning (RC-Align).}
To achieve fine-grained alignment between images and text, we explore the matching between image regions and medical textual concepts. 

Let the $j$-th concept in the $n$-th text be $G_{n,j}$, with the concept token indices in the original text represented by $\{v_c\}_{c=1}^U$, where $(1\leq v_c\leq s)$ and $U$ is the length of the concept, i.e., the number of words within the concept. The corresponding concept representation is then given by:
\begin{equation}
    \mathbf{g}_{n,j}=\text{mean\_pooling}(\mathbf{t}_{(n,v_1)}, \mathbf{t}_{(n,v_2)},...,\mathbf{t}_{(n,v_U)}).
\end{equation}

For the $m$-th image and the $n$-th text, we define a region-concept similarity matrix $\mathbf{A}\in\mathbb{R}^{r\times w}$, where $w$ represents the number of concepts in the $n$-th text. The alignment score $\mathbf{a}_{ij}$ for the $i$-th region in the $m$-th image and the $j$-th concept in the $n$-th text is calculated as:
\begin{equation}\mathbf{a}_{ij}=\frac{(\mathbf{g}_{n,j})^T\mathbf{i}_{m,i}}{||\mathbf{g}_{n,j}||\cdot ||\mathbf{i}_{m,i}||}.
\end{equation}

The similarity matrix element $\mathbf{A}_{ij}$ is then expressed as:
\begin{equation}
    \mathbf{A}_{ij}=\text{log}\frac{1}{1+e^{-t_l\mathbf{a}_{i,j}+b_l}}.
    \label{eq:rc-align}
\end{equation}

The parameters $t_l$ and $b_l$ are the logit scale and logit bias for the local region-concept alignment, respectively.

For the $m$-th image $I_m$ and the $n$-th text $T_n$, the similarity score is:
\begin{equation}
    S(I_m,T_n)=\frac{1}{w}\sum_{j=1}^{w}\max_i(\mathbf{A}_{ij}).
\end{equation}

Thus, the RC-Align loss in a mini-batch of $|\mathcal{B}|$ image-text-concept triplets is:
\begin{equation}
    \mathcal{L}_{\text{RC-Align}}=-\frac{1}{|\mathcal{B}|}\sum_{m=1}^{|\mathcal{B}|}\sum_{n=1}^{|\mathcal{B}|}z^{m,n}\cdot S(I_m,T_n).
\end{equation}

This ensures that the representations of the most similar image region and textual concept are closer when the image and the text are paired, and further apart otherwise.

Overall, the total training loss is:
\begin{equation}
\mathcal{L}=\mathcal{L}_{\text{IT-Align}}+\alpha\mathcal{L}_{\text{RC-Align}},
\label{eq:train-loss}
\end{equation}
where $\alpha$ is a hyper-parameter to adjust the weight of the RC-Align loss during pre-training.

\subsubsection*{Inference with local and global information integration}
\label{sec:inference_local}

For zero-shot classification, our model integrates global and local alignment through dual-path scoring. Given an image \( I_m \) with global representation \( \mathbf{i}_{m,\text{cls}} \in \mathbb{R}^h \) and local region features \( \mathbf{i}_m = [\mathbf{i}_{m,1}, ..., \mathbf{i}_{m,r}] \in \mathbb{R}^{r \times h} \), let class \( c \) have global text embedding \( \mathbf{t}_{c,\text{cls}} \in \mathbb{R}^h \) and \( w_c \) concept embeddings \( \mathbf{g}_{c} = [\mathbf{g}_{c,1}, ..., \mathbf{g}_{c,w_c}] \in \mathbb{R}^{w_c \times h} \).

\textbf{Global scoring} computes similarity between normalized global representations:
\begin{equation}
    s_g(c) = t_g \cdot \frac{{\mathbf{i}_{m,\text{cls}}^\top \mathbf{t}_{c,\text{cls}}}}{{\|\mathbf{i}_{m,\text{cls}}\| \|\mathbf{t}_{c,\text{cls}}\|}} + b_g.
    \label{eq:gscore-revised}
\end{equation}

\textbf{Local scoring} aggregates region-concept information through pooled representations. Different from what we have done in the proposed RC-Align, we take a region selection criterion based on top-$k$ regions per concept:
\begin{equation}
\begin{aligned}
\mathbf{i}_{m,\text{loc}} &= \frac{1}{w_c k} \sum_{j=1}^{w_c} \sum_{i \in \mathcal{T}_{c,j}^k} \mathbf{i}_{m,i}, \quad
    \mathbf{t}_{c,\text{loc}} = \frac{1}{w_c}\sum_{j=1}^{w_c} \mathbf{g}_{c,j}, \\
    s_l(c) &= t_l \cdot \frac{{\mathbf{i}_{m,\text{loc}}^\top \mathbf{t}_{c,\text{loc}}}}{{\|\mathbf{i}_{m,\text{loc}}\| \|\mathbf{t}_{c,\text{loc}}\|}} + b_l,
    \label{eq:lscore-revised}
\end{aligned}
\end{equation}

where the region selection criterion operates as follows:
\[
\forall i \in \mathcal{T}_{c,j}^k, \forall n \notin \mathcal{T}_{c,j}^k: \frac{\mathbf{i}_{m,i}^\top \mathbf{g}_{c,j}}{\|\mathbf{i}_{m,i}\| \|\mathbf{g}_{c,j}\|} \geq \frac{\mathbf{i}_{m,n}^\top \mathbf{g}_{c,j}}{\|\mathbf{i}_{m,n}\| \|\mathbf{g}_{c,j}\|}.
\]

Here, $\mathcal{T}_{c,j}^k \subset \{1, \dots, r\}$ denotes the set of indices for the top-$k$ regions most similar to concept $j$ in class $c$. 

\textbf{Probability fusion} combines both pathways:
\begin{equation}
\begin{aligned}
    p_g(c) &= \frac{e^{s_g(c)}}{\sum_{d=1}^C e^{s_g(d)}}, \quad 
    p_l(c) = \frac{e^{s_l(c)}}{\sum_{d=1}^C e^{s_l(d)}}, \\
    p(c) &= \beta p_l(c) + (1-\beta) p_g(c).
    \label{eq:prob-fusion-revised}
\end{aligned}
\end{equation}

The final prediction \( \hat{c} = \arg\max_c p(c) \) integrates both alignment strategies through learnable parameters \( t_g, b_g, t_l, b_l \) from Equations \ref{eq:it-align} and \ref{eq:rc-align}, with \( \beta \) controlling the fusion ratio.

\subsection*{Experimental settings}
In this section, we first describe the dataset used to pre-train the proposed \ours{} model. Based on the pre-trained model, we conduct extensive experiments on downstream tasks. Besides, we describe the state-of-the-art medical multi-modal foundation models used for detailed comparison. Finally, we cover the implementation details in pre-training the proposed \ours{}.

\subsubsection*{Pre-training dataset}
In our pre-training process, the proposed \ourd{} is used. In detail, it contains 23 million image-text-concept triplets, where the image-text pairs are collected from PMC-OA and concepts are aligned to the UMLS concepts and extracted from the text.

\subsubsection*{Implementation details}
We develop \ours{} by utilizing SigLIP-ViT-400M-16\footnote{\url{https://huggingface.co/timm/ViT-SO400M-14-SigLIP-384}} as the visual encoder and PubMedBERT\footnote{\url{https://huggingface.co/microsoft/BiomedNLP-BiomedBERT-base-uncased-abstract- fulltext}} as the text encoder. 
Each input image is resized to $336 \times 336$ pixels. During pre-training, we employ the AdamW optimizer~\cite{Loshchilov2017DecoupledWD} with a learning rate of 0.0005. 
The model is initially pre-trained on 48 H800 GPUs with a batch size of 12,288 for 32 epochs on the \ourd{} dataset, without the RC-Align loss.
Subsequently, we 
conduct further pre-training with the RC-Align loss on this dataset with a batch size of 6,144 and learning rate of 0.0003 for 20 epochs. The weight $\alpha$ (in Equation \ref{eq:train-loss}) during the pre-training is set to 0.5.

In medical image diagnosis, for zero-shot classification, we set $\beta$ (in Equation \ref{eq:prob-fusion-revised}) to 0.5 so that the prediction can consider the information from both global information and local information. 
For linear probing, we use the LogisticRegression function provided in scikit-learn\footnote{\url{https://scikit-learn.org/1.5/modules/generated/sklearn.linear_model.LogisticRegression.html}}, the inverse of regularization strength is 0.316, maximum iteration is 1,000, random state is 1, and all other parameters are default. For all fully fine-tuning experiments, the batch size is 64, epochs is 20, learning rate is 0.0001. We use AdamW optimizer for training the model.
In the visual question answering task, we use AdamW as the optimizer. For SigLIP-400M and \ours{}, the batch size is 32, learning rate is 0.000001. For other models, the batch size is 4, learning rate is 0.000005. All experiments have the same epoch number of 50.
In pathology whole-slide image analysis, the batch size is 1, the learning rate is 0.0002, the epoch number is 30, and the cosine learning rate scheduler is used.

\subsubsection*{Downstream tasks}

\textbf{Medical image diagnosis.}
We assess the zero-shot, linear probe (using 1\%, 10\%, and 100\% of the training data), and fully fine-tuning (using 100\% of the training data) capabilities of \ours{} and previous state-of-the-art medical CLIP models across 36 datasets (\textbf{Extended Data Table \ref{tab:datasets}}) spanning 10 image modalities. In the zero-shot setting, no ground truth labels are provided. We follow the zero-shot protocol of the CLIP model~\cite{radford2021learning}, where the text encoder and image encoder are used to encode the medical images and labels (using a prompt template to form a sentence), respectively, and we then calculate the similarity and assign labels to the images. For linear probing, we first extract features from each image using a specific image encoder from a pre-trained model and employ these image features and their labels to perform logistic regression. The trained logistic regression models are subsequently used to test model performance. This experiment aims to evaluate each model's ability to extract discriminative image features. In the fully fine-tuning setting, each model is concatenated with a classification head and fine-tuned on a specific dataset with all parameters adjustable. We use the AUC score for model evaluation to ensure that the performance is not influenced by thresholds.

\textbf{Cross-modal image-text retrieval.}
Cross-modal image-text retrieval includes tasks such as medical image-to-text and text-to-image retrieval. To evaluate the performance of \ours{} across various medical image modalities, we utilize two datasets: PMC-9K and Quilt-1M~\cite{ikezogwo2024quilt}, to assess the models' cross-modal retrieval capabilities in a mix of image modalities and pathology images. We use Image-to-Text Recall@1,5,10 and Text-to-Image Recall@1,5,10 to evaluate each model's performance.

\textbf{Medical report generation.}
Medical report generation aims to alleviate the workload of doctors by automatically generating a report given a medical image. In our study, following the shallow alignment in the R2GenGPT framework~\cite{wang2023r2gengpt}, we replace the vision encoder with that of each medical CLIP model and use the same training scheme to fine-tune the report generation model. We conduct these experiments on the MIMIC-CXR~\cite{johnson2019mimic} and IU X-Ray~\cite{demner2016preparing} datasets. To comprehensively evaluate each model's performance, we use natural language generation metrics: BLEU-1,2,3,4, METEOR, ROUGE-L, CIDEr, and clinical efficacy metrics: Micro Precision, Micro Recall, and Micro F1.

\textbf{Medical visual question answering.}
\label{sec:method_vqa}
Medical Visual Question Answering (VQA) serves as a bridge between AI systems and humans. Given a medical image and a question about the image, the model provides an answer to the question based on the image. In our experiments, we use the METER framework~\cite{dou2022empirical} to evaluate different models on the VQA task, employing the SLAKE~\cite{liu2021slake} and VQA-RAD~\cite{lau2018dataset} datasets. Following METER's evaluation metrics, we use Closed Accuracy, Open Accuracy, and Overall Accuracy to assess each model's performance.

\textbf{Pathology whole-slide image analysis.}
To explore our model's ability to understand pathology whole-slide images, we conduct experiments on 6 related datasets for 9 subtasks in cancer diagnosis, mutation prediction, and survival prediction, each of which can be regarded as a classification task, with a whole-slide image as input to the vision encoder and a predicted label as output. Given the high resolution of whole-slide images, we follow the conventional approach by splitting the whole-slide image into smaller regions and using each pre-trained vision encoder to extract features from each image region. These features are input into the ABMIL model~\cite{ilse2018attention} for multiple instance learning (MIL). The model outputs a predicted label. Similar to the medical image diagnosis task, we use the AUC score to evaluate different models in cancer diagnosis and molecular subtyping, and the concordance index (C-index) for survival prediction.

\textbf{Medical concept annotation.}
We evaluate to what extent the proposed \ours{} can annotate fine-grained concepts for medical images, hence facilitating the development of concept-based explainable artificial intelligence. The performance of medical concept annotation is evaluated using the concept labels of the datasets. For each concept, two positive and negative prompts are designed to calculate the similarity with input images to get the predicted output. Specifically, we adopt five datasets with clinical concept annotations to evaluate the zero-shot concept annotation performance, including dermatology images, blood cell images (pathology), breast ultrasound, and chest CT.

\textbf{Inherently interpretable model employment and evaluation.}
Building inherently interpretable models is essential for deploying AI in medicine since it is more trustworthy for doctors and patients during clinical diagnosis. In this paper, we employ the Concept Bottleneck Model (CBM) \cite{cbm} to build the inherently interpretable model based on our method. Inspired by previous works \cite{labo,label_free_cbm}, we use ConceptCLIP to calculate the cosine similarity between input images and medical concepts, then the similarity is used as the input to the concept bottleneck layer, which maps the concept similarity to the final prediction. The concept bottleneck layer is trained using the class labels. For concept-class association, the weights of each neuron within the linear layer can be regarded as the contribution to the final prediction. To evaluate the effectiveness of our method, we compare ConceptCLIP with black-box models and other vision-language models of both the natural and medical domains on four datasets of different modalities, including dermatology, pathology, ultrasound, and CT.

\textbf{Discovery for correlations between concepts and diseases.}
To conduct correlation discovery between concepts and diseases, concept presence difference is calculated for datasets of four modalities, including NLM-TB (chest X-ray), CBIS-DDSM (mammography), UBIBC (breast ultrasound), and MedFMC-Colon (pathology). Specifically, for each dataset, assume there are two image sets of different classes (e.g., \textit{tuberculosis} and \textit{normal} in NLM-TB dataset), denoted as $I_+ = \{I_{1+}, I_{2+}, ..., I_{M+}\}$ and $I_- = \{I_{1-}, I_{2-}, ..., I_{N-}\}$, and the pre-defined concept list $C$ (e.g., \textit{cavitation, consolidation}, etc.), where $M$ and $N$ are the numbers of images in the positive and negative image sets, respectively, and $C$ are generated by a large language model (i.e., GPT-4 \cite{gpt4}) since the dataset does not have concept labels. We adopt the proposed ConceptCLIP to annotate the presence of the concepts for each given image. The concept presence proportion of each image set is computed by $\frac{N_{c_i}}{M}$ and $\frac{N_{c_i}}{N}$, respectively, here $N_{c_i}$ is the number of images where a specific concept $c_i$ is present. Then the concept presence difference of concept $c_i$ is defined as the difference between the two concept presence proportions, i.e., $D_{c_i} = \frac{N_{c_i}}{M} - \frac{N_{c_i}}{N}$. A positive value of $D_{c_i}$ means concept $c_i$ is more likely to appear in the positive image set $I_+$, and vice versa.
For disease-level concept inspection, as shown in \textbf{Fig. \ref{fig:xai_combined} (i)}, pneumonia is selected as the inspected disease. For concept inspection on a dataset from one institution, the RSNA \cite{shih2019augmenting} dataset is adopted, which only includes classes of \textit{No finding} and \textit{Pneumonia}. For disease-level concept inspection, we collect datasets from various sources that also include the two classes, i.e., VinDr-CXR from the Hospital 108 and the Hanoi Medical University Hospital \cite{nguyen2022vindr} and VinDr-PCXR from the Phu Tho Obstetric \& Pediatric Hospital \cite{pham2022vindr}, where VinDr-PCXR focuses on the chest X-rays of children. It is noteworthy that VinDr-CXR and VinDr-PCXR contain other classes such as \textit{Lung tumor} and \textit{Tuberculosis}, but we only consider the images of \textit{No Finding} and \textit{Pneumonia} to analyze the pneumonia disease. For each dataset, we use ConceptCLIP to obtain the concept-disease association. Then we average the weights of the concept bottleneck layer across all considered datasets to derive the final top relevant concepts for pneumonia.

\subsubsection*{Metrics}

\textbf{AUC.}
The Area Under the Receiver Operating Characteristic Curve (AUC) is a performance measurement for classification problems at various threshold settings. It is widely used in medical image diagnosis and explainable AI tasks to evaluate the ability of a model to distinguish between classes.

\textbf{Accuracy.}  
Accuracy (ACC) quantifies the proportion of correct predictions relative to the total evaluated cases.   

\[
\text{ACC} = \frac{\text{Number of Correct Predictions}}{\text{Total Number of Predictions}}.
\]
For visual question answering (VQA) tasks, we evaluate three variants: \textbf{closed accuracy}, which measures performance on predefined answer candidates (e.g., multiple-choice questions); \textbf{open accuracy}, assessing free-form answer generation without predefined options; and \textbf{overall accuracy}, providing a combined assessment of both closed and open settings.

\textbf{C-Index.}
The concordance index (C-Index) measures the predictive accuracy of a survival model. It is particularly used in survival prediction tasks in whole-slide image analysis to evaluate how well the model predicts the order of events. It is calculated as the proportion of all usable patient pairs whose predictions and outcomes are concordant.

\textbf{Recall.}  
Recall (sensitivity) quantifies the proportion of relevant instances successfully retrieved from the total relevant population. In classification tasks, this metric evaluates a model’s ability to minimize false negatives (FN), defined as:  

\[
\text{Recall} = \frac{\mathrm{TP}}{\mathrm{TP} + \mathrm{FN}},  
\]  
where \(\mathrm{TP}\) (true positives) represents correct identifications of positive-class instances, and \(\mathrm{FN}\) (false negatives) denotes missed positive instances.  
For \textbf{cross-modal retrieval} (e.g., image-to-text or text-to-image search), we evaluate performance using \(\text{Recall}@k\), which measures the proportion of queries where relevant items appear in the top-\(k\) retrieved results. Specifically, \(\text{Image-to-Text Recall}@k\) represents the fraction of image queries for which the corresponding ground-truth text caption appears among the top-\(k\) matches, while \(\text{Text-to-Image Recall}@k\) measures the analogous proportion for text queries retrieving their paired images. Higher \(\text{Recall}@k\) values (typically computed for \(k = 1, 5, 10\)) indicate more robust cross-modal alignment in the model's learned representations.

\textbf{BLEU.}  
The Bilingual Evaluation Understudy (BLEU) score~\cite{papineni2002bleu} quantifies the similarity between machine-generated text and reference human-authored texts by measuring lexical and syntactic overlap. Originally developed for machine translation, BLEU is widely adopted in medical report generation to evaluate both semantic fidelity (alignment with clinical content) and fluency (linguistic coherence).  
The score is computed as:  
\[
\text{BLEU} = \underbrace{\text{BP}}_{\text{Brevity Penalty}} \cdot \exp\left(\sum_{n=1}^{N} w_n \log p_n\right),  
\]  
where \(p_n\) represents the modified n-gram precision (the fraction of machine-generated n-grams that match the reference texts), \(w_n\) are weighting factors (typically \(w_n = \frac{1}{N}\) for n-grams up to length \(N=4\)), and BP is the brevity penalty \(\min\left(1, e^{1 - \frac{\text{reference length}}{\text{generated length}}}\right)\) that prevents artificially high scores for shorter outputs.
Higher BLEU scores (range: 0–1) indicate stronger alignment with reference texts. In medical applications, this metric ensures generated reports preserve critical clinical terminology (via \(n\)-gram precision) while maintaining natural narrative structure (via brevity penalty).

\textbf{METEOR.}
The Metric for Evaluation of Translation with Explicit ORdering (METEOR)~\cite{banerjee2005meteor} is another metric for evaluating text generation quality, focusing on precision, recall, and alignment of phrases. It is used for medical report generation to provide a more nuanced assessment of generated text quality.

\textbf{ROUGE-L.}
The Recall-Oriented Understudy for Gisting Evaluation (ROUGE-L)~\cite{lin2004rouge} measures the longest common subsequence (LCS) between the generated and reference text sequences. It is used for evaluating medical report generation to capture the overlap in content. The formula for ROUGE-L is:
\[
\text{ROUGE-L} = \frac{LCS(X, Y)}{\text{length of reference}}.
\]

\textbf{CIDEr.}
Consensus-based Image Description Evaluation (CIDEr)~\cite{vedantam2015cider} is a metric designed to assess the similarity of a generated text to multiple reference text sequences, emphasizing consensus. It is used for medical report generation to evaluate the relevance and quality of the generated content.

\textbf{Clinical Efficacy.}  
Clinical efficacy assesses the practical utility of AI-generated medical reports by measuring their concordance with clinical standards and potential impact on diagnostic decisions. We employ micro-averaged precision, recall, and F1 scores computed using the CheXbert classifier~\cite{irvin2019chexpert}, which identifies 14 standardized radiological observations (including pneumothorax and edema) in both generated and reference reports.
The micro-averaged approach aggregates performance across all disease classes: \textbf{Micro Precision} measures the fraction of correctly identified findings among all system-predicted abnormalities (true positives/[true positives + false positives]), while \textbf{Micro Recall} calculates the proportion of reference report findings correctly detected by the system (true positives/[true positives + false negatives]). The \textbf{Micro F1} score, defined as:
\[
\text{Micro F1} = 2 \cdot \frac{\text{Micro Precision} \cdot \text{Micro Recall}}{\text{Micro Precision} + \text{Micro Recall}},
\]
provides a balanced measure of diagnostic performance by combining precision and recall. This comprehensive evaluation framework ensures clinically meaningful assessment of report quality, where high precision minimizes overdiagnosis and high recall reduces missed findings.
These metrics reflect system-wide reliability: high precision minimizes overdiagnosis (false positives), while high recall mitigates missed diagnoses (false negatives). Micro-averaging prioritizes per-case accuracy over per-class balancing, mirroring clinical workflows where all observations in a report hold equal urgency.

\subsection*{Data availability}
The pre-training dataset, described in Section \ref{sec:method}, is derived from publicly available sources and will be released to the research community upon publication to ensure reproducibility. All evaluation datasets referenced in \textbf{Extended Data Table \ref{tab:datasets}} are publicly accessible benchmarks, with complete citation information provided to enable direct access by other researchers.

\subsection*{Code availability}
The implementation of \ours{} is available at \url{https://github.com/JerrryNie/ConceptCLIP}. The weights of \ours{} is released at \url{https://huggingface.co/JerrryNie/ConceptCLIP}.

\subsection*{Author contribution}
Y.N., S.H., Y.B., and H.C. conceived and designed the work. Y.N., S.H., Y.B., Y.W., Zhixuan C., Zhiyuan C., and H.W. collected the data for pre-training and downstream task evaluation. Y.N. and S.H. contributed to the technical implementation of the pre-training process. Y.B. contributed to the technical implementation of explainable AI. Y.N. and S.H. evaluated medical image diagnosis tasks. Y.N. evaluated cross-modal retrieval tasks. Y.B. evaluated explainable AI tasks. Y.W. evaluated whole-slide image analysis tasks. Zhixuan C. evaluated medical report generation tasks. S.Y. evaluated medical visual question answering tasks. Y.N., Y.B., and S.H. wrote the manuscript with inputs from all authors. 
Xi.W., L.L., M.W., Xian.W., R.C.K.C., Y.M.L., Y.Z., and P.R. provided suggestions on the framework and experiments.
All authors reviewed and approved the final paper. H.C. supervised the research. 

\subsection*{Declarations}
The authors have no conflicts of interest to declare.

\subsection*{Ethics declarations}
This study has been reviewed and approved by the Human and Artefacts Research Ethics Committee (HAREC). The protocol number is HREP-2024-0212.

\subsection*{Acknowledgements}
This work was supported by the Hong Kong Innovation and Technology Commission (Project No. MHP/002/22, GHP/006/22GD and ITCPD/17-9), HKUST (Project No. FS111), and the Research Grants Council of the Hong Kong Special Administrative Region, China (Project Reference Number: T45-401/22-N),

\bibliography{sn-bibliography}
\newpage
\begin{appendices}
\section{Extended Data}
\fontsize{8}{11}\selectfont
\begin{longtable}{|p{.15\textwidth}|p{.70\textwidth}|p{.05\textwidth}|}
\caption{Medical image analysis datasets grouped by task. Click the hyperlink in the ``link'' column to access the corresponding dataset.}\label{tab:datasets} \\
\toprule
Dataset & Description & Link \\
\midrule
\endfirsthead
\multicolumn{3}{c}{{\bfseries \tablename\ \thetable{} -- continued from previous page}} \\
\toprule
Dataset & Description & Link \\
\midrule
\endhead
\midrule
\multicolumn{3}{r}{{Continued on next page}} \\
\bottomrule
\endfoot
\bottomrule
\endlastfoot
\multicolumn{3}{|c|}{\textbf{Binary Classification}} \\
\midrule
SIIM-ACR~\cite{siim-acr-pneumothorax-segmentation} & A binary classification dataset containing chest radiographic
images. & \href{https://www.kaggle.com/c/siim-acr-pneumothorax-segmentation}{\checkmark} \\
\midrule
Covid-CXR2~\cite{Wang2020} & A dataset of over 16,000 chest X-ray images from more than 15,100 patients across 51 countries, including 2,300 positive COVID-19 images, designed to differentiate between no pneumonia, non-COVID-19 pneumonia, and COVID-19 pneumonia in the COVIDx V8A dataset, while the COVIDx V8B dataset focuses on detecting COVID-19 positive and negative cases. & \href{https://www.kaggle.com/datasets/andyczhao/covidx-cxr2}{\checkmark} \\
\midrule
NLM-TB~\cite{jaeger2014two} & A radiology image dataset with 800 total images with the label of normal or tuberculosis. & \href{https://www.ncbi.nlm.nih.gov/pmc/articles/PMC4256233/}{\checkmark} \\
\midrule
LC25000 (Colon)~\cite{borkowski2019lung} & A dataset comprises 10,000 pathology images from colon tissues, featuring benign tissues and adenocarcinomas. & \href{https://github.com/tampapath/lung_colon_image_set}{\checkmark} \\
\midrule
UBIBC~\cite{ubibc2023} & A dataset consists of ultrasound images related to breast cancer,
including benign and malignant. & \href{https://www.kaggle.com/datasets/vuppalaadithyasairam/ultrasound-breast-images-for-breast-cancer}{\checkmark} \\
\midrule
PCam200~\cite{kawai2023large} & A dataset of pathological H\&E images created from the Camelyon2016 challenge dataset~\cite{bejnordi2017diagnostic}. & \href{https://github.com/enigmanx20/patchtcga}{\checkmark} \\
\midrule
RSNA~\cite{shih2019augmenting} & A dataset of about 30,000 frontal view chest X-ray images, each labeled for binary classification to indicate the presence of pneumonia. & \href{https://www.rsna.org/rsnai/ai-image-challenge/rsna-pneumonia-detection-challenge-2018}{\checkmark} \\
\midrule
Brain Tumor CT~\cite{likhon2023brain} & A dataset of CT scans for brain tumor detection and analysis. It features high-resolution images from multiple patients, labeled with tumor types (e.g., glioma, meningioma) and their locations in the brain, designed to aid in developing AI models for automatic detection, classification, and segmentation of brain tumors. & \href{https://www.kaggle.com/datasets/murtozalikhon/brain-tumor-multimodal-image-ct-and-mri}{\checkmark} \\
\midrule
Brain Tumor MRI~\cite{likhon2023brain} & This dataset is similar to the Brain Tumor CT dataset~\cite{likhon2023brain} but with the MRI modality. & \href{https://www.kaggle.com/datasets/murtozalikhon/brain-tumor-multimodal-image-ct-and-mri}{\checkmark} \\
\midrule
DDSM~\cite{skooch2023ddsm} & A dataset containing 55,890 pre-processed images from the DDSM~\cite{heath2001digital} and CBIS-DDSM~\cite{lee2016curated} datasets, formatted as 299$\times$299 pixel TFRecords for TensorFlow, with 14\% positive and 86\% negative examples, split into test and validation sets. & \href{https://www.kaggle.com/datasets/skooch/ddsm-mammography}{\checkmark} \\
\midrule
Breast Cancer~\cite{hayder_2024} & A dataset of 3,383 annotated mammogram images focused on breast tumors, exported from Roboflow, designed for building and testing deep learning models for tumor detection. & \href{https://www.kaggle.com/datasets/hayder17/breast-cancer-detection/}{\checkmark} \\
\midrule
HMC-QU~\cite{degerli2024early} & A dataset created by Hamad Medical Corporation, Tampere University, and Qatar University includes 2D echocardiography recordings from 2018 and 2019 for heart attack detection and left ventricle wall segmentation, with recordings at 25 fps and resolutions from 422$\times$636 to 768$\times$1024 pixels. & \href{https://www.kaggle.com/datasets/aysendegerli/hmcqu-dataset/data}{\checkmark} \\
\midrule
\multicolumn{3}{|c|}{\textbf{Multi-Label Classification}} \\
\midrule
RFMiD2~\cite{panchal2023retinal} & A multi-label dataset of fundus images
annotated by three eye specialists. & \href{https://www.mdpi.com/2306-5729/8/2/29}{\checkmark} \\
\midrule
MedFMC (Colon)~\cite{wang2023real} & A dataset focused on facilitating the detection of early-stage cancer cells in tissue slides, enabling pathologists to classify and quantify cancerous regions. & \href{https://github.com/openmedlab/MedFM}{\checkmark} \\
\midrule
VinDr-Mammo~\cite{nguyen2023vindr} & A large-scale benchmark dataset for computer-aided detection and diagnosis in full-field digital mammography. & \href{https://vindr.ai/datasets/mammo}{\checkmark} \\
\midrule
VinDr-PCXR~\cite{pham2022vindr} & A dataset focuses on pediatric chest X-rays for the interpretation of common thoracic diseases. & \href{https://vindr.ai/datasets/pediatric-chest-x-ray}{\checkmark} \\
\midrule
VinDr-CXR~\cite{nguyen2022vindr} & A dataset of chest X-Ray images containing 18,000 high-quality postero-anterior scans with detailed localization of 22 critical findings and classification of 6 thoracic diseases, annotated by experienced radiologists from major hospitals in Vietnam. 
& \href{https://vindr.ai/datasets/cxr}{\checkmark} \\
\midrule
VinDr-SpineXR~\cite{nguyen2021vindr} & A dataset of annotated medical images for detecting and classifying spinal lesions from radiographs. & \href{https://vindr.ai/datasets/spinexr}{\checkmark} \\
\midrule
NIH ChestX-ray14~\cite{wang2017chestx} & A dataset comprising 112,120 frontal-view X-ray images from 30,805 unique patients, featuring fourteen disease labels identified through natural language processing (NLP) techniques, collected between 1992 and 2015. & \href{https://nihcc.app.box.com/v/ChestXray-NIHCC}{\checkmark} \\
\midrule
CheXpert~\cite{irvin2019chexpert} & A dataset of chest radiographs from 65,240 patients. & \href{https://stanfordmlgroup.github.io/competitions/chexpert/}{\checkmark} \\
\midrule
\multicolumn{3}{|c|}{\textbf{Multi-Class Classification}} \\
\midrule
DRD~\cite{diabetic-retinopathy-detection} & A dataset consists of high-resolution retina images labeled by subject ID and eye side, with each image rated for diabetic retinopathy severity on a scale of 0 (no DR) to 4 (proliferative DR) by ophthalmologists. & \href{https://www.kaggle.com/competitions/diabetic-retinopathy-detection}{\checkmark} \\
\midrule
LC25000 (Lung)~\cite{borkowski2019lung} & A dataset consists of 15,000 pathology images from lung tissue, including benign tissues and various carcinomas. & \href{https://github.com/tampapath/lung_colon_image_set}{\checkmark} \\
\midrule
MedFMC (Chest)~\cite{wang2023real} & A dataset for screening thoracic diseases encompasses 19 common thoracic abnormalities. & \href{https://github.com/openmedlab/MedFM}{\checkmark} \\
\midrule
MedFMC (Endo)~\cite{wang2023real} & A dataset aimed at enhancing the automatic detection and classification of four different lesion types in colonoscopy images, facilitating early diagnosis of colorectal cancer. & \href{https://github.com/openmedlab/MedFM}{\checkmark} \\
\midrule
HAM10000~\cite{tschandl2018ham10000} & A dataset consists of dermatoscopic images of pigmented lesions. & \href{https://www.kaggle.com/datasets/kmader/skin-cancer-mnist-ham10000}{\checkmark} \\
\midrule
BUSBRA~\cite{gomez2024bus} & A dataset of anonymized breast ultrasound images from 1,064 patients, including biopsy-proven tumors, BI-RADS annotations, and ground truth delineations of tumoral and normal regions. & \href{https://zenodo.org/records/8231412}{\checkmark} \\
\midrule
WCE~\cite{WCE} & A dataset of colon disease images containing curated samples for training and testing, derived from the Kvasir~\cite{pogorelov2017kvasir} and ETIS-Larib-Polyp DB~\cite{silva2014toward} datasets. & \href{https://www.kaggle.com/datasets/francismon/curated-colon-dataset-for-deep-learning}{\checkmark} \\
\midrule
Fundus JSIEC~\cite{cen2021automatic} & A dataset of 1,000 fundus images belonging to 39 classes, sourced from the Joint Shantou International Eye Centre in Shantou city, Guangdong province, China. & \href{https://www.kaggle.com/datasets/linchundan/fundusimage1000}{\checkmark} \\
\midrule
HyperKvasir~\cite{Borgli2020} & A dataset consisting of 10,662 labeled images in the JPEG format, categorized into 23 classes of medical findings, highlighting the imbalance in the number of images per class. & \href{https://datasets.simula.no/hyper-kvasir/}{\checkmark} \\
\midrule
Kvasir~\cite{pogorelov2017kvasir} & A dataset of annotated and verified gastrointestinal tract images, featuring various classes of anatomical landmarks and pathological findings, with resolutions ranging from 720$\times$576 to 1920$\times$1072 pixels. & \href{https://datasets.simula.no/kvasir/}{\checkmark} \\
\midrule
ODIR~\cite{li2021benchmark} & A dataset of 5,000 patients featuring color fundus photographs from both left and right eyes, along with ophthalmologists' diagnostic keywords, with 6,392 samples used for training. & \href{https://www.kaggle.com/datasets/andrewmvd/ocular-disease-recognition-odir5k}{\checkmark} \\
\midrule
BUID~\cite{al2020dataset} & A dataset of 780 breast ultrasound images from 600 female patients aged 25 to 75, collected in 2018, categorized into three classes: normal, benign, and malignant. & \href{https://www.kaggle.com/datasets/aryashah2k/breast-ultrasound-images-dataset}{\checkmark} \\
\midrule
PAD-UFES-20~\cite{pacheco2020pad} & A dataset of skin lesion images labeled into six categories: basal cell carcinoma, squamous cell carcinoma, actinic keratosis, seborrheic keratosis, melanoma, and nevus. & \href{https://www.notion.so/c1e0ff8f36814b94b205ee919e2e3ff8?pvs=21}{\checkmark} \\
\midrule
OCTMNIST~\cite{medmnistv2} & A dataset for multi-class classification of retinal OCT images. & \href{https://medmnist.com/}{\checkmark} \\
\midrule
Breast Tumor MRI~\cite{msoud_nickparvar_2021} & A combined dataset (including three datasets: figshare~\cite{Cheng2017}, SARTAJ dataset~\cite{sartaj2020}. and Br35H dataset~\cite{hamada2020br35h}) of 7,023 human brain MRI images classified into four classes: glioma, meningioma, no tumor, and pituitary. The ``no tumor'' class images are from the Br35H dataset. & \href{https://www.kaggle.com/datasets/masoudnickparvar/brain-tumor-mri-dataset}{\checkmark} \\
\midrule
Retinal OCT~\cite{subramanian2022classification} & A dataset of high-quality retinal OCT images categorized into 8 classes of retinal diseases. & \href{https://www.kaggle.com/datasets/obulisainaren/retinal-oct-c8}{\checkmark} \\
\midrule
COVIDxCT~\cite{Gunraj2022} & A dataset of CT scans consisting of two variants, ``A'' with confirmed COVID-19 cases and ``B'' which includes weakly verified cases. We choose the ``A'' set in our experiment. & \href{https://www.kaggle.com/datasets/hgunraj/covidxct/data}{\checkmark} \\
\midrule
\multicolumn{3}{|c|}{\textbf{Retrieval}} \\
\midrule
PMC-9K &  newly curated dataset consisting of a cleaned, held-out collection of
9,222 image-text pairs in \ourd{}. It is designed to comprehensively evaluate
the cross-modal retrieval capabilities of \ours{} across various modalities. & - \\
\midrule
Quilt-1M~\cite{ikezogwo2024quilt} & A large-scale histopathology collection containing 768,826 image-text pairs. For evaluation purposes, we filter out the data sourced from PubMed Central and utilize a held-out subset comprising pathology image-text pairs to assess the retrieval performance. & \href{https://quilt1m.github.io/}{\checkmark} \\
\midrule
\multicolumn{3}{|c|}{\textbf{Medical Report Generation}} \\
\midrule
MIMIC-CXR~\cite{johnson2019mimic} & A dataset contains 371,920 chest X-rays linked to 227,943 imaging studies from 65,079 patients. & \href{https://physionet.org/content/mimic-cxr/2.1.0/}{\checkmark} \\
\midrule
IU X-Ray~\cite{demner2016preparing} & A dataset consists of 7,470 pairs of chest X-ray images and their corresponding diagnostic reports. & \href{https://www.kaggle.com/datasets/raddar/chest-xrays-indiana-university/data}{\checkmark} \\
\midrule
\multicolumn{3}{|c|}{\textbf{Visual Question Answering}} \\
\midrule
VQA-RAD~\cite{lau2018dataset} & A dataset manually constructed, featuring 3,064 question-answer pairs where clinicians ask questions about radiology images and provide reference answers. & \href{https://osf.io/89kps/}{\checkmark} \\
\midrule
SLAKE~\cite{liu2021slake} & A bilingual radiology VQA dataset that includes 642 images and 14,000 questions. We only use the English part. & \href{https://www.med-vqa.com/slake/}{\checkmark} \\
\midrule
\multicolumn{3}{|c|}{\textbf{Cancer Diagnosis}} \\
\midrule
BRACS-3~\cite{brancati2022bracs} & A dataset contains 6 different subtypes of lesions, including images representing atypical lesions. Histological images representing normal tissue samples are also included. In this setting, a breast tumor according to a pathology image can be classified into ``benign'', ``atypical'', or ``malignant''. & \href{https://www.bracs.icar.cnr.it/}{\checkmark} \\
\midrule
BRACS-7~\cite{brancati2022bracs} & A dataset that contains the same images as BRACS-3 but with a different label set. In this setting, a breast tumor according to a pathology image can be classified into ``normal'', ``pathological benign'', ``usual ductal hyperplasia'', ``flat epithelial atypia'', ``atypical ductal hyperplasia'', ``ductal carcinoma in situ'', or ``invasive carcinoma''. & \href{https://www.bracs.icar.cnr.it/}{\checkmark} \\
\midrule
NSCLC~\cite{weinstein2013cancer} & A dataset contains 1,053 histopathology slides of non-small cell lung cancer, including 541 lung adenocarcinoma and 512 lung squamous cell carcinoma cases. & \href{https://www.cancer.gov/ccg/research/genome-sequencing/tcga}{\checkmark} \\
\midrule
Camelyon~\cite{bejnordi2017diagnostic,litjens20181399} & A dataset based on CAMELYON16~\cite{bejnordi2017diagnostic} and CAMELYON17~\cite{litjens20181399}, which evaluates new and existing algorithms for automated detection and classification of breast cancer metastases in whole-slide images of histological lymph node sections. & \href{https://camelyon16.grand-challenge.org/https://camelyon17.grand-challenge.org/}{\checkmark} \\
\midrule
\multicolumn{3}{|c|}{\textbf{Mutation Prediction}} \\
\midrule
BRCA-TP53~\cite{weinstein2013cancer} & This dataset comprises molecular and clinical data from the Breast Invasive Carcinoma (BRCA) cohort of The Cancer Genome Atlas (TCGA), with specialized annotation for TP53-mutant tumors, the most frequently altered tumor suppressor in this malignancy. & \href{https://www.cancer.gov/ccg/research/genome-sequencing/tcga}{\checkmark} \\
\midrule
BRCA-TNN~\cite{weinstein2013cancer} & This dataset comprises molecular and clinical profiles of triple-negative breast tumors (TNBC) from the TCGA BRCA cohort, defined by immunohistochemical absence of estrogen receptor (ER), progesterone receptor (PR), and HER2 amplification. & \href{https://www.cancer.gov/ccg/research/genome-sequencing/tcga}{\checkmark} \\
\midrule
LUAD-TP53~\cite{weinstein2013cancer} & This dataset provides comprehensive molecular profiling of TP53-mutant lung adenocarcinoma cases from the TCGA cohort. & \href{https://www.cancer.gov/ccg/research/genome-sequencing/tcga}{\checkmark} \\
\midrule
\multicolumn{3}{|c|}{\textbf{Survival Prediction}} \\
\midrule
BRCA~\cite{weinstein2013cancer} & A TCGA dataset of breast carcinoma for survival prediction.& \href{https://www.cancer.gov/ccg/research/genome-sequencing/tcga}{\checkmark} \\
\midrule
LUSC~\cite{weinstein2013cancer} & A dataset of the Lung Squamous Cell Carcinoma LUSC cohort from the TCGA for survival prediction.& \href{https://www.cancer.gov/ccg/research/genome-sequencing/tcga}{\checkmark} \\
\midrule
\end{longtable}
\fontsize{8}{11}\selectfont\begin{longtable}{|p{2cm}|p{7cm}|p{3cm}|}
\caption{Zero-shot experimental settings for medical image analysis datasets.} \label{tab:zero_shot_settings} \\
\toprule
\textbf{Dataset} & \textbf{Classes} & \textbf{Prompts} \\
\midrule
\endfirsthead
\multicolumn{3}{c}{{\bfseries \tablename\ \thetable{} -- continued from previous page}} \\
\toprule
\textbf{Dataset} & \textbf{Classes} & \textbf{Prompts} \\
\midrule
\endhead
\midrule
\multicolumn{3}{r}{{Continued on next page}} \\
\endfoot
\bottomrule
\endlastfoot
SIIM-ACR~\cite{siim-acr-pneumothorax-segmentation} & No Finding, Pneumothorax & \texttt{a chest radiology presents \{\}} \\
\midrule
RFMiD2~\cite{panchal2023retinal} & Neovascularization, Macular Edema, Myopia, Retinal Traction, Coloboma, Choroidal Folds, Tortuous Vessels, Retinitis Pigmentosa, Retinal Pigment Epithelium Changes, Optic Disc Pallor, Media Haze, Retinitis, Preretinal Hemorrhage, Asteroid Hyalosis, Drusens, Hemorrhagic Pigment Epithelial Detachment, Branch Retinal Vein Occlusion, Optic Disc Edema, Exudation, Haemorrhagic Retinopathy, Tilted Disc, Tessellation, Retinal Tears, Retinal Detachment, Optic Disc Cupping, Macular Hole, Silicone Oil-Filled Eye, Cotton Wool Spots, Vasculitis, Microaneurysm, Macular Scar, Age-Related Macular Degeneration, Optic Neuritis, Anterior Ischemic Optic Neuropathy, Laser Scar, Chorioretinitis, Within Normal Limit, Epiretinal Membrane, Central Retinal Vein Occlusion, Central Serous Retinopathy, Optociliary Shunt & \texttt{an image of \{\}} \\
\midrule
DRD~\cite{diabetic-retinopathy-detection} & No Diabetic Eetinopathy, Mild Diabetic Retinopathy, Moderate Diabetic Retinopathy, Severe Diabetic Retinopathy, Proliferative Diabetic Retinopathy & \texttt{a detailed view of a retina indicating \{\}}; \newline\texttt{a close-up of a retina highlighting \{\}} \\
\midrule
COVID-CXR2~\cite{Wang2020} & No Finding, COVID-19 & \texttt{a radiographic representation assessing for \{\}} \\
\midrule
NLM-TB~\cite{jaeger2014two} & Normal, Tuberculosis & \texttt{a close-up view of a chest x-ray presenting evidence of \{\}}; \newline\texttt{a visual analysis of a chest x-ray with signs of \{\}}; \newline\texttt{a radiographic scan highlighting the presence of \{\}}; \newline\texttt{an annotated image from a chest x-ray showing signs of \{\}} \\
\midrule
LC25000 (Colon)~\cite{borkowski2019lung} & Normal Colonic Tissue, Colon Adenocarcinomas & \texttt{the histopathological image illustrates that of \{\}}; \newline\texttt{a histopathological image featuring \{\} in colon}; \newline\texttt{a pathological image highlighting a \{\}} \\
\midrule
LC25000 (Lung)~\cite{borkowski2019lung} & Lung Squamous Cell carcinomas, Lung Adenocarcinomas, Normal Lung Tissue & \texttt{histopathological image contains \{\}}; \newline\texttt{a pathological image highlighting a \{\}}; \newline\texttt{an annotated histopathological image representing a \{\}} \\
\midrule
MedFMC (Chest)~\cite{wang2023real} & Pleural Effusion, Nodule, Pneumonia, Cardiomegaly, Hilar Enlargement, Fracture Old, Fibrosis, Aortic Calcification, Tortuous Aorta, Thickened Pleura, TB, Pneumothorax, Emphysema, Atelectasis, Calcification, Pulmonary Edema, Increased Lung Markings, Elevated Diaphragm, Consolidation & \texttt{lung situation of \{\}} \\
\midrule
MedFMC (Colon)~\cite{wang2023real} & Tumor, Normal & \texttt{this slide features an annotated section indicating a \{\}}; \newline\texttt{the image illustrates a \{\} in the context of colon tissue} \\
\midrule
MedFMC (Endo)~\cite{wang2023real} & Ulcer, Erosion, Polyp, Tumor & \texttt{a diagnostic endoscopy showing features of \{\}}; \newline\texttt{an endoscopic finding suggestive of \{\}} \\
\midrule
VinDr-Mammo~\cite{nguyen2023vindr} & Birads Negative, Breast Heterogeneously Density, No Finding, Breast Scattered Areas of Fibroglandular, Birads Suspicious Malignant, Mass, Breast Extremely Density, Birads Highly Suggestive of Malignant, Suspicious Calcification, Birads Benign, Breast Almost Entirely Fatty, Suspicious Lymph Node, Focal Asymmetry, Birads Probably Benign, Asymmetry, Architectural Distortion, Skin Thickening, Global Asymmetry, Nipple Retraction, Skin Retraction & \texttt{an image of \{\}} \\
\midrule
HAM10000~\cite{tschandl2018ham10000} & Actinic Keratoses and Intraepithelial Carcinoma, Basal Cell Carcinoma, Benign Keratosis-Like Lesions, Dermatofibroma, Melanoma, Melanocytic Nevi, Vascular Lesions & \texttt{a dermatology has that of \{\} presented} \\
\midrule
VinDr-PCXR~\cite{pham2022vindr} & Bronchiolitis, Pneumonia, Other Disease, Bronchitis, Brocho-Pneumonia, Tuberculosis, Mediastinal Tumor, Nan, Situs Inversus, Hyaline Membrane Disease, CPAM, No Finding & \texttt{an image of \{\}} \\
\midrule
BUSBRA~\cite{gomez2024bus} & Metaplasia Apocrina, Ductal Carcinoma in Situ, Lipoma, Papillary Carcinoma, Hyperplasia, Hamartoma, Ductal Hyperplasia, Cyst, Lymphoma, Invasive Ductal Carcinoma, Fibroadenoma, Lymph Node, Sclerosing Adenosis, Medullary Carcinoma, Adenocarcinoma, Intraductal Papilloma, Duct Ectasia, Mastitis, Fibrosis, Phyllodes Tumor, Lobular Atrophy, Lobular Carcinoma in Situ, Fibrocystic Changes, Mucinous Carcinoma, Galactocele, Invasive Lobular Carcinoma, Proliferative Lesions, Fat Necrosis & \texttt{this is an image of \{\}}; \newline\texttt{\{\} presented in image} \\
\midrule
VinDr-CXR~\cite{nguyen2022vindr} & Atelectasis, Lung Tumor, Lung Cavity, Tuberculosis, Pulmonary Fibrosis, Clavicle Fracture, Lung Cyst, Pneumonia, Calcification, No Finding, Rib Fracture, Pleural Thickening, Other Disease, Mediastinal Shift, Enlarged PA, Nodule/Mass, ILD, COPD, Pneumothorax, Consolidation, Infiltration, Pleural Effusion, Other Lesion, Cardiomegaly, Emphysema, Lung Opacity & \texttt{an image of \{\}} \\
\midrule
UBIBC~\cite{ubibc2023} & Benign, Malignant & \texttt{\{\} show in an ultrasound image of breast cancer}; \newline\texttt{an ultrasound image of breast cancer showing signs of \{\}} \\
\midrule
WCE~\cite{WCE} & Normal, Ulcerative Colitis, Polyps, Esophagitis & \texttt{an endoscopic finding suggestive of \{\}} \\
\midrule
PCam200~\cite{kawai2023large} & Normal, Tumor & \texttt{histopathological image contains \{\}}; \newline\texttt{an annotated histopathological image representing a \{\}} \\
\midrule
Fundus JSIEC~\cite{cen2021automatic} & Massive Hard Exudates, Vitreous Particles, Possible Glaucoma, Yellow-White Spots-Flecks, CRVO, BRVO, Large Optic Cup, Blur Fundus without PDR, Bietti Crystalline Dystrophy, Rhegmatogenous RD, CSCR, Fibrosis, Congenital Disc Abnormality, Laser Spots, Vessel Tortuosity, Maculopathy, RAO, Pathological Myopia, DR2, Tessellated Fundus, Retinitis Pigmentosa, ERM, VKH Disease, MH, Cotton-Wool Spots, Optic Atrophy, Severe Hypertensive Retinopathy, DR3, DR1, Preretinal Hemorrhage, Fundus Neoplasm, Silicon Oil in Eye, Blur Fundus with Suspected PDR, Myelinated Nerve Fiber, Normal, Disc Swelling and Elevation, Peripheral Retinal Degeneration and Break, Chorioretinal Atrophy-Coloboma, Dragged Disc & \texttt{\{\} presented in image} \\
\midrule
HyperKvasir~\cite{Borgli2020} & Barrett's, Barrett's Short Segment, BBPS 0-1, BBPS 2-3, Cecum, Dyed Lifted Polyps, Dyed Resection Margins, Esophagitis A, Esophagitis B-D, Hemorrhoids, Terminal Ileum, Impacted Stool, Polyps, Pylorus, Retroflex Rectum, Retroflex Stomach, Ulcerative Colitis 0-1, Ulcerative Colitis 1, Ulcerative Colitis 1-2, Ulcerative Colitis 2, Ulcerative Colitis 2-3, Ulcerative Colitis 3, Z-Line & \texttt{this is an image of \{\}}; \newline\texttt{\{\} presented in image} \\
\midrule
Kvasir~\cite{pogorelov2017kvasir} & Dyed Lifted Polyps, Dyed Resection Margins, Esophagitis, Normal Cecum, Normal Pylorus, Normal Z Line, Polyps, Ulcerative Colitis & \texttt{an endoscopic finding suggestive of \{\}}; \newline\texttt{a snapshot from an endoscopic procedure showing \{\}} \\
\midrule
RSNA~\cite{shih2019augmenting} & No Finding, Pneumonia & \texttt{the chest radiology image of \{\}}; \newline\texttt{the lung radiology illustrates that of \{\}} \\
\midrule
VinDr-SpineXR~\cite{nguyen2021vindr} & No Finding, Osteophytes, Disc Space Narrowing, Surgical Implant, Foraminal Stenosis, Other Lesions, Vertebral Collapse, Spondylolisthesis & \texttt{a spine x-ray image of \{\}} \\
\midrule
NIH ChestX-ray14~\cite{wang2017chestx} & Atelectasis, Cardiomegaly, Effusion, Infiltration, Mass, Nodule, Pneumonia, Pneumothorax, Consolidation, Edema, Emphysema, Fibrosis, Pleural Thickening, Hernia, No Finding & \texttt{an image of \{\}} \\
\midrule
CheXpert~\cite{irvin2019chexpert} & Enlarged Cardiomediastinum, Cardiomegaly, Lung Opacity, Lung Lesion, Edema, Consolidation, Pneumonia, Atelectasis, Pneumothorax, Pleural Effusion, Pleural Other, Fracture, Support Devices, No Finding & \texttt{an image of \{\}} \\
\midrule
ODIR~\cite{li2021benchmark} & Normal, Diabetes, Glaucoma, Cataract, Age Related Macular Degeneration, Hypertension, Pathological Myopia, Other Diseases/Abnormalities & \texttt{\{\} presented in retinography} \\
\midrule
BUID~\cite{al2020dataset} & Normal, Malignant, Benign & \texttt{a close-up ultrasound scan showing a \{\} in the breast} \\
\midrule
PAD-UFES-20~\cite{pacheco2020pad} & Basal Cell Carcinoma, Squamous Cell Carcinoma, Actinic Keratosis, Seborrheic Keratosis, Melanoma, Nevus & \texttt{a dermatological image of \{\}}; \newline\texttt{a diagnostic dermatological image illustrating a \{\}} \\
\midrule
OCTMNIST~\cite{medmnistv2} & Choroidal Neovascularization, Diabetic Macular Edema, Drusen, Normal & \texttt{an image of \{\}} \\
\midrule
Breast Tumor MRI~\cite{msoud_nickparvar_2021} & Glioma, Meningioma, Pituitary, No Tumor & \texttt{a breast mri image of \{\}} \\
\midrule
Retinal OCT~\cite{subramanian2022classification} & Age-Related Macular Degeneration, Choroidal Neovascularization, Central Serous Retinopathy, Diabetic Macular Edema, Macular Hole, Drusen, Diabetic Retinopathy, Normal & \texttt{a retinal oct image of \{\}} \\
\midrule
Brain Tumor CT~\cite{likhon2023brain} & Healthy, Tumor & \texttt{an image of \{\}} \\
\midrule
COVIDxCT~\cite{Gunraj2022} & Normal, Pneumonia, COVID-19 & \texttt{an image of \{\}} \\
\midrule
Brain Tumor MRI~\cite{likhon2023brain} & Healthy, Tumor & \texttt{an image of \{\}} \\
\midrule
DDSM~\cite{skooch2023ddsm} & Negative, Positive & \texttt{the mammography image of \{\}} \\
\midrule
Breast Cancer~\cite{hayder_2024} & Normal, Tumor & \texttt{the mammography image of \{\}} \\
\midrule
HMC-QU~\cite{degerli2024early} & Apical 4 Chamber, Apical 2 Chamber & \texttt{an image of \{\}} \\
\midrule
\end{longtable}

\begin{figure*}[t] 
\centering 
\includegraphics[width=\linewidth]{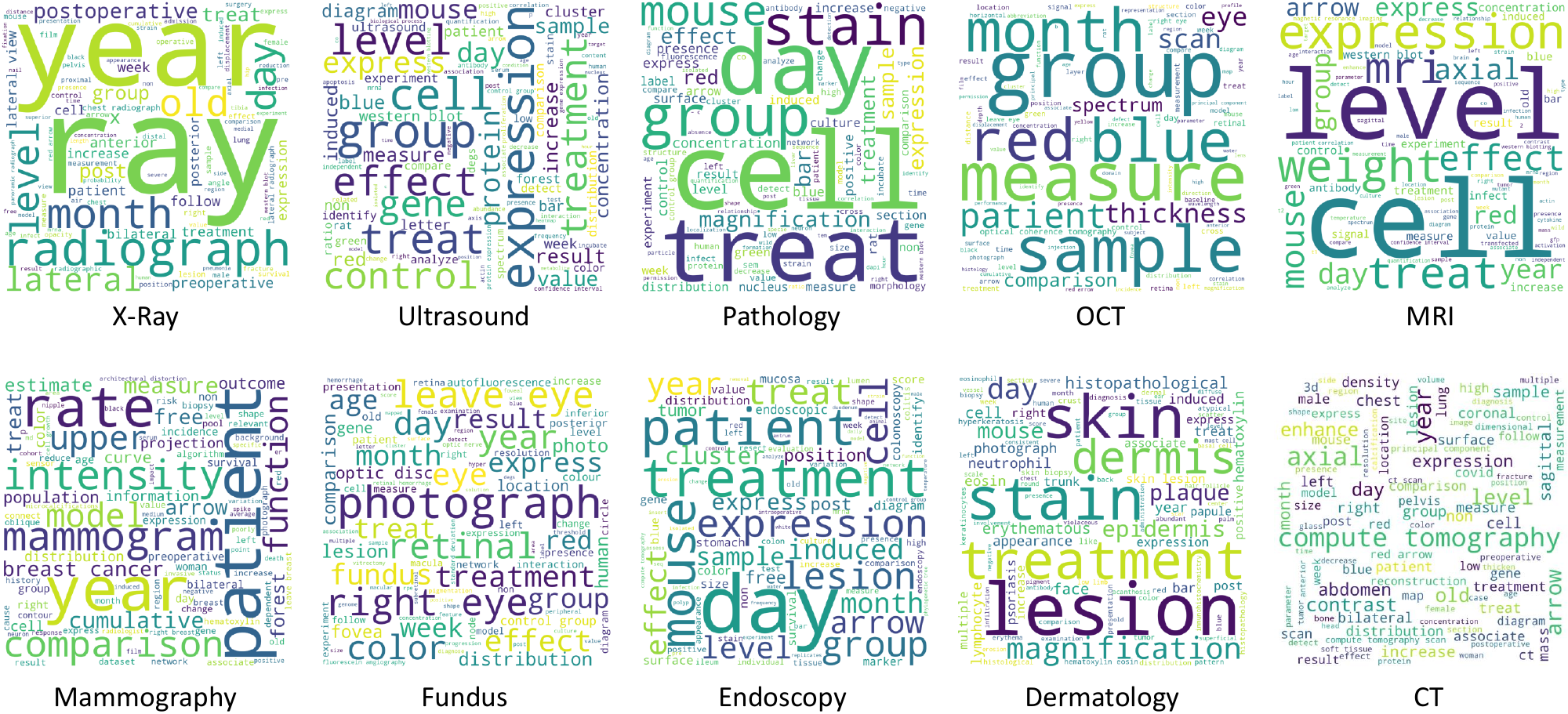} 
\caption{ 
\textbf{Concepts from the \ourd{} dataset.} 
Wordclouds of concepts in captions to qualitatively visualize the concepts in each medical image modality. Word size is proportional to its occurrence in \ourd{}. Common nouns and verbs are ignored.} \label{fig:wordcloud_by_modality} 
\end{figure*} 
\clearpage
\begin{table*}[h]
\centering
\caption{{AUC scores (\%) for classification results across different modalities in the zero-shot setting. \textbf{Bold} indicates the best results and \underline{underline} indicates the second best.The 95\% CI is included in parentheses.}}
\resizebox{\textwidth}{!}{%
\begin{tabular}{l|c|c|c|c|c}
\toprule
Dataset & CLIP & SigLIP-400M & PMC-CLIP & BiomedCLIP & ConceptCLIP \\ 
\midrule
\multicolumn{6}{c}{\textbf{X-Ray}} \\ 
\midrule
SIIM-ACR & 55.13 (50.86, 59.26) & 67.87 (64.13, 71.31) & 64.19 (60.43, 67.68) & \underline{77.08} (73.96, 80.15) & \textbf{83.05} (80.40, 85.63) \\ 
Covid-CXR2 & 49.40 (47.53, 51.26) & \underline{70.34} (68.70, 72.16) & 57.41 (55.26, 59.32) & 68.99 (66.88, 70.98) & \textbf{81.77} (80.19, 83.35) \\ 
NLM-TB & 65.64 (55.98, 73.96) & 82.92 (76.45, 88.80) & 74.22 (65.71, 81.75) & \underline{88.60} (82.44, 94.08) & \textbf{92.92} (88.48, 96.69) \\ 
MedFMC (Chest) & 49.81 (47.37, 50.00) & \underline{61.80} (57.86, 64.99) & 49.85 (47.37, 50.00) & 50.20 (47.62, 50.70) & \textbf{66.54} (62.12, 70.67) \\ 
VinDr-PCXR & 42.40 (35.27, 48.47) & 43.38 (35.73, 48.83) & \underline{45.69} (36.56, 51.25) & \textbf{48.02} (38.63, 55.35) & 43.72 (34.59, 49.95) \\ 
VinDr-CXR & 48.92 (45.78, 50.09) & \underline{58.58} (53.25, 61.85) & 49.06 (46.05, 50.01) & 48.84 (45.46, 50.03) & \textbf{61.24} (55.13, 65.68) \\
\bottomrule
\end{tabular}
}
\label{tab:combined_zero_shot_cls_scores}
\end{table*}

\begin{table*}[h]
\centering
\caption{{AUC scores (\%) for classification results across different modalities in the linear probing setting with 1\% training data. \textbf{Bold} indicates the best results and \underline{underline} indicates the second best. The 95\% CI is included in parentheses.}}
\resizebox{\textwidth}{!}{%
\begin{tabular}{l|c|c|c|c|c}
\toprule
Dataset & CLIP & SigLIP-400M & PMC-CLIP & BiomedCLIP & ConceptCLIP \\ 
\midrule
\multicolumn{6}{c}{\textbf{X-Ray}} \\ 
\midrule
SIIM-ACR & 69.96 (66.47, 73.58) & \underline{79.29} (76.32, 82.24) & 74.66 (71.40, 77.90) & 73.04 (69.62, 76.06) & \textbf{88.38} (86.15, 90.42) \\ 
Covid-CXR2 & \textbf{96.29} (95.73, 96.84) & \underline{93.39} (92.62, 94.07) & 90.56 (89.65, 91.43) & 88.27 (87.12, 89.40) & 87.52 (86.42, 88.68) \\ 
NLM-TB & \textbf{75.80} (67.51, 83.60) & 65.80 (56.69, 74.00) & 62.22 (53.17, 71.13) & 54.10 (44.33, 62.72) & \underline{68.15} (59.59, 76.48) \\ 
MedFMC (Chest) & 53.49 (50.90, 55.39) & 56.52 (53.81, 58.49) & 60.81 (58.13, 62.55) & \underline{61.28} (58.21, 63.09) & \textbf{62.94} (60.42, 64.54) \\ 
VinDr-PCXR & 46.80 (39.07, 52.96) & 47.32 (39.54, 53.54) & 47.24 (39.63, 53.31) & \textbf{48.01} (40.18, 53.93) & \underline{47.74} (40.15, 53.86) \\ 
VinDr-CXR & 52.42 (49.12, 54.31) & 63.31 (59.90, 65.25) & 58.91 (55.17, 60.90) & \underline{64.35} (60.68, 66.07) & \textbf{68.39} (65.06, 70.13) \\
\bottomrule
\end{tabular}
}
\label{tab:combined_linear_probes_1_cls_scores}
\end{table*}

\begin{table*}[h]
\centering
\caption{{AUC scores (\%) for classification results across different modalities in the linear probing setting with 10\% training data. \textbf{Bold} indicates the best results and \underline{underline} indicates the second best. The 95\% CI is included in parentheses.}}
\resizebox{\textwidth}{!}{%
\begin{tabular}{l|c|c|c|c|c}
\toprule
Dataset & CLIP & SigLIP-400M & PMC-CLIP & BiomedCLIP & ConceptCLIP \\ 
\midrule
\multicolumn{6}{c}{\textbf{X-Ray}} \\ 
\midrule
SIIM-ACR & 71.58 (68.30, 74.92) & 80.05 (76.99, 82.91) & 78.56 (75.52, 81.38) & \underline{81.24} (78.42, 83.90) & \textbf{88.73} (86.56, 90.78) \\ 
Covid-CXR2 & \textbf{97.29} (96.80, 97.76) & \underline{96.44} (95.93, 96.93) & 95.07 (94.46, 95.66) & 92.61 (91.75, 93.46) & 92.61 (91.77, 93.37) \\ 
NLM-TB & 71.82 (63.16, 79.77) & 86.91 (81.10, 91.73) & 79.61 (72.06, 86.45) & \underline{87.02} (81.32, 92.16) & \textbf{95.11} (91.00, 98.03) \\ 
MedFMC (Chest) & 59.73 (55.85, 62.79) & 67.63 (63.81, 70.43) & 71.55 (68.03, 74.18) & \underline{72.82} (69.32, 75.39) & \textbf{73.87} (70.22, 77.31) \\ 
VinDr-PCXR & \textbf{51.10} (41.11, 59.05) & 47.69 (39.61, 52.36) & \underline{50.72} (42.99, 56.39) & 50.30 (41.55, 56.65) & 47.97 (40.30, 52.75) \\ 
VinDr-CXR & 64.66 (59.72, 67.22) & 73.96 (68.81, 76.30) & 71.84 (66.61, 74.25) & \underline{74.59} (69.22, 76.80) & \textbf{78.26} (73.81, 81.06) \\
\bottomrule
\end{tabular}
}
\label{tab:combined_linear_probes_10_cls_scores}
\end{table*}

\begin{table*}[h]
\centering
\caption{{AUC scores (\%) for classification results across different modalities in the linear probing setting with 100\% training data. \textbf{Bold} indicates the best results and \underline{underline} indicates the second best. The 95\% CI is included in parentheses.}}
\resizebox{\textwidth}{!}{%
\begin{tabular}{l|c|c|c|c|c}
\toprule
Dataset & CLIP & SigLIP-400M & PMC-CLIP & BiomedCLIP & ConceptCLIP \\ 
\midrule
\multicolumn{6}{c}{\textbf{X-Ray}} \\ 
\midrule
SIIM-ACR & 74.43 (70.89, 77.61) & 83.33 (80.66, 86.03) & 83.45 (80.62, 86.06) & \underline{85.49} (83.02, 87.95) & \textbf{90.12} (88.08, 92.20) \\ 
Covid-CXR2 & \underline{98.24} (97.87, 98.63) & \textbf{98.74} (98.46, 99.00) & 98.11 (97.75, 98.45) & 96.07 (95.50, 96.60) & 96.94 (96.46, 97.37) \\ 
NLM-TB & 70.02 (61.77, 78.29) & 87.38 (81.15, 92.84) & 79.54 (71.57, 86.50) & \underline{88.69} (83.13, 93.07) & \textbf{95.05} (91.50, 98.18) \\ 
MedFMC (Chest) & 66.28 (62.32, 69.67) & 73.40 (68.56, 76.44) & \underline{76.33} (72.02, 78.66) & 76.31 (72.24, 79.53) & \textbf{79.40} (75.04, 82.54) \\ 
VinDr-PCXR & 61.39 (48.30, 70.05) & 63.77 (50.39, 72.24) & \textbf{67.28} (53.24, 78.07) & 63.59 (49.15, 71.75) & \underline{63.89} (51.40, 72.75) \\ 
VinDr-CXR & 69.67 (63.36, 73.04) & 77.36 (71.45, 81.09) & 77.74 (72.06, 81.40) & \underline{79.45} (73.26, 82.37) & \textbf{82.72} (75.55, 86.44) \\
\bottomrule
\end{tabular}
}
\label{tab:combined_linear_probes_100_cls_scores}
\end{table*}
 
\begin{table*}[h]
\centering
\caption{{AUC scores (\%) for classification results across different modalities in the fully fine-tuning setting. \textbf{Bold} indicates the best results and \underline{underline} indicates the second best. The 95\% CI is included in parentheses.}}
\resizebox{\textwidth}{!}{%
\begin{tabular}{l|c|c|c|c|c}
\toprule
Dataset & CLIP & SigLIP-400M & PMC-CLIP & BiomedCLIP & ConceptCLIP \\ 
\midrule
\multicolumn{6}{c}{\textbf{X-Ray}} \\ 
\midrule
SIIM-ACR & 88.29 (86.07, 90.45) & 89.46 (87.19, 91.55) & 90.05 (87.94, 92.09) & \underline{90.59} (88.56, 92.47) & \textbf{91.72} (89.82, 93.58) \\ 
Covid-CXR2 & 99.76 (99.53, 99.91) & \textbf{99.93} (99.85, 99.98) & 99.76 (99.57, 99.91) & 99.85 (99.66, 99.97) & \underline{99.89} (99.77, 99.98) \\ 
NLM-TB & 71.37 (63.30, 79.07) & \underline{92.83} (88.02, 97.04) & 87.07 (81.66, 92.69) & 92.64 (87.29, 97.12) & \textbf{94.32} (90.11, 97.65) \\ 
MedFMC (Chest) & \underline{75.70} (71.89, 78.92) & 73.56 (69.91, 76.59) & 72.50 (68.61, 75.99) & 74.41 (70.19, 78.33) & \textbf{82.41} (78.08, 84.86) \\ 
VinDr-PCXR & 57.89 (46.23, 66.38) & \underline{65.14} (50.66, 74.33) & 59.53 (47.55, 68.18) & 60.69 (49.10, 70.10) & \textbf{67.25} (53.99, 76.13) \\ 
VinDr-CXR & 62.09 (56.20, 67.83) & 70.30 (64.01, 74.03) & 58.00 (53.27, 62.36) & \underline{71.43} (65.68, 75.29) & \textbf{83.47} (77.23, 86.20) \\
\bottomrule
\end{tabular}
}
\label{tab:combined_fully_finetune_cls_scores}
\end{table*}
 
\begin{table*}[h]
\centering
\caption{Recall@1, 5, 10 (\%) of different models for cross-modal retrieval on the PMC-9K dataset. \textbf{Bold} indicates the best results and \underline{underline} indicates the second best. The 95\% CI is included in parentheses.}
\begin{tabular}{l|ccc}
\hline
\multicolumn{4}{c}{\textbf{Image-to-Text}} \\ \hline
Model & Recall@1 & Recall@5 & Recall@10 \\ \hline
CLIP & 5.63 (5.18, 6.10) & 12.47 (11.80, 13.15) & 17.14 (16.38, 17.92) \\ 
SigLIP-400M & 2.86 (2.54, 3.23) & 7.04 (6.53, 7.59) & 9.68 (9.09, 10.37) \\ 
PMC-CLIP & 15.77 (15.05, 16.49) & 34.33 (33.34, 35.37) & 44.35 (43.35, 45.35) \\ 
BiomedCLIP & \underline{73.41} (72.53, 74.32) & \underline{91.93} (91.40, 92.46) & \underline{95.30} (94.88, 95.75) \\ 
ConceptCLIP & \textbf{82.85} (82.05, 83.60) & \textbf{94.71} (94.23, 95.20) & \textbf{97.01} (96.61, 97.38) \\ 
\hline
\multicolumn{4}{c}{\textbf{Text-to-Image}} \\ \hline
Model & Recall@1 & Recall@5 & Recall@10 \\ \hline
CLIP & 6.31 (5.79, 6.83) & 13.11 (12.44, 13.86) & 18.39 (17.63, 19.24) \\ 
SigLIP-400M & 2.65 (2.32, 2.97) & 6.96 (6.47, 7.47) & 9.82 (9.22, 10.40) \\ 
PMC-CLIP & 15.69 (14.96, 16.43) & 32.93 (31.96, 33.82) & 42.71 (41.67, 43.69) \\ 
BiomedCLIP & \underline{74.02} (73.15, 74.93) & \underline{92.06} (91.49, 92.65) & \underline{95.38} (94.96, 95.80) \\ 
ConceptCLIP & \textbf{83.24} (82.51, 83.96) & \textbf{94.50} (94.05, 94.99) & \textbf{96.86} (96.50, 97.21) \\ 
\hline
\end{tabular}
\label{tab:retrieval_metrics_pmc}
\end{table*}

\begin{table*}[h]
\centering
\caption{Recall@1, 50, 200 (\%) of different models for cross-modal retrieval on the QUILT-1M dataset. \textbf{Bold} indicates the best results and \underline{underline} indicates the second best. The 95\% CI is included in parentheses.}
\begin{tabular}{l|ccc}
\hline
\multicolumn{4}{c}{\textbf{Image-to-Text}} \\ \hline
Model & Recall@1 & Recall@50 & Recall@200 \\ \hline
CLIP & 0.62 (0.42, 0.58) & 4.14 (3.88, 4.56) & 8.37 (7.95, 8.84) \\ 
SigLIP-400M & 0.28 (0.19, 0.30) & 3.82 (3.53, 4.15) & 7.98 (7.57, 8.46) \\ 
PMC-CLIP & 0.31 (0.21, 0.33) & 4.87 (4.53, 5.27) & 13.32 (12.66, 13.88) \\ 
BiomedCLIP & \underline{0.68} (0.46, 0.65) & \underline{8.36} (7.80, 8.80) & \underline{18.88} (18.12, 19.47) \\ 
ConceptCLIP & \textbf{1.86} (1.31, 1.57) & \textbf{15.13} (14.51, 15.72) & \textbf{28.70} (27.91, 29.60) \\ 
\hline
\multicolumn{4}{c}{\textbf{Text-to-Image}} \\ \hline
Model & Recall@1 & Recall@50 & Recall@200 \\ \hline
CLIP & \underline{0.74} (0.48, 0.65) & 5.18 (4.80, 5.49) & 9.40 (8.92, 9.90) \\
SigLIP-400M & 0.36 (0.22, 0.35) & 4.05 (3.72, 4.39) & 9.24 (8.76, 9.80) \\ 
PMC-CLIP & 0.18 (0.14, 0.25) & 5.23 (4.84, 5.62) & 13.36 (12.72, 13.90) \\ 
BiomedCLIP & 0.74 (0.48, 0.66) & \underline{9.14} (8.56, 9.55) & \underline{19.62} (18.91, 20.31) \\ 
ConceptCLIP & \textbf{1.95} (1.38, 1.67) & \textbf{17.07} (16.44, 17.73) & \textbf{32.50} (31.79, 33.45) \\ 
\hline
\end{tabular}
\label{tab:retrieval_metrics_quilt}
\end{table*} 
\begin{table*}[h]
\centering
\caption{{Results (\%) on the medical visual question answering task.  \textbf{Bold} indicates the best results and \underline{underline} indicates the second best. The 95\% CI is included in parentheses.}}
\resizebox{\textwidth}{!}{%
\begin{tabular}{l|c|c|c|c|c}
\toprule
Metric & CLIP & BiomedCLIP & PMC-CLIP & SigLIP-400M & ConceptCLIP \\ 
\midrule
\multicolumn{6}{c}{\textbf{SLAKE}} \\ 
\midrule
Closed Accuracy& 83.87 (80.29, 87.26) & 84.86 (81.49, 88.22) & 83.80 (80.04, 87.50) & \underline{85.60} (84.85, 90.88) & \textbf{87.97} (84.85, 90.88) \\ 
Open Accuracy& 77.24 (73.95, 80.47) & \underline{80.70} (78.29, 84.34) & 79.09 (76.12, 82.33) & 77.49 (74.11, 80.62) & \textbf{81.27} (78.29, 84.34) \\ 
Overall Accuracy& 79.78 (77.29, 82.28) & \underline{82.31} (81.53, 85.96) & 81.03 (78.51, 83.13) & 80.69 (78.42, 83.03) & \textbf{83.86} (81.53, 85.96) \\ 
\midrule
\multicolumn{6}{c}{\textbf{VQA-RAD}} \\ 
\midrule
Closed Accuracy& 80.04 (74.90, 84.86) & 79.70 (74.50, 84.46) & \textbf{82.47} (76.89, 86.45) & 78.85 (73.31, 84.06) & \underline{81.72} (76.89, 86.45) \\ 
Open Accuracy& 54.39 (47.50, 61.00) & 54.08 (47.50, 61.50) & \underline{55.70} (49.50, 64.00) & 51.50 (44.50, 58.00) & \textbf{56.97} (49.50, 64.00) \\ 
Overall Accuracy& 68.71 (64.52, 72.73) & 68.21 (63.63, 72.51) & \underline{70.50} (66.73, 74.73) & 66.73 (62.53, 70.96) & \textbf{70.70} (66.73, 74.73) \\ 
\midrule
\bottomrule
\end{tabular}
}
\label{tab:vqa_scores}
\end{table*}
\begin{table*}[h]
\centering
\caption{{Results (\%) on the medical report generation task.  \textbf{Bold} indicates the best results and \underline{underline} indicates the second best. The 95\% CI is included in parentheses.}}
\resizebox{\textwidth}{!}{%
\begin{tabular}{l|c|c|c|c|c}
\toprule
Metric & CLIP & BiomedCLIP & PMC-CLIP & SigLIP-400M & ConceptCLIP \\ 
\midrule
\multicolumn{6}{c}{\textbf{MIMIC-CXR}} \\ 
\midrule
BLEU-1& \underline{36.20} (36.13, 36.73) & 35.84 (35.54, 36.15) & 32.87 (32.60, 33.12) & 34.76 (34.48, 35.05) & \textbf{36.45} (36.13, 36.73) \\ 
BLEU-2& \underline{21.70} (21.65, 22.25) & 21.64 (21.35, 21.96) & 17.87 (17.61, 18.12) & 20.82 (20.54, 21.11) & \textbf{21.95} (21.65, 22.25) \\ 
BLEU-3& \underline{13.41} (13.32, 13.94) & 13.38 (13.07, 13.70) & 10.00 (9.75, 10.24) & 12.56 (12.24, 12.86) & \textbf{13.63} (13.32, 13.94) \\ 
BLEU-4& 8.28 (7.99, 8.59) & \underline{8.31} (8.06, 8.67) & 5.57 (5.37, 5.79) & 7.69 (7.42, 7.96) & \textbf{8.36} (8.06, 8.67) \\ 
ROUGE-L& \underline{26.89} (26.95, 27.51) & 26.48 (26.21, 26.76) & 23.27 (23.05, 23.48) & 25.95 (25.71, 26.22) & \textbf{27.21} (26.95, 27.51) \\ 
METEOR& \underline{15.41} (15.49, 15.84) & 15.19 (15.03, 15.36) & 14.02 (13.88, 14.18) & 14.19 (14.03, 14.36) & \textbf{15.66} (15.49, 15.84) \\
\bottomrule
\end{tabular}
}
\label{tab:mrg_scores}
\end{table*}
\begin{table*}[h!]
\centering
\caption{Results (\%) on various datasets for pathology whole-slide image tasks. We use the C-index for survival prediction tasks. For other tasks, we use AUC scores. \textbf{Bold} indicates the best results and \underline{underline} indicates the second best. The 95\% CI is included in parentheses.}
\resizebox{\textwidth}{!}{
\begin{tabular}{l|c|c|c|c|c}
\toprule
Dataset & CLIP & BiomedCLIP & PMC-CLIP & SigLIP-400M  & ConceptCLIP \\
\midrule
\multicolumn{6}{c}{\textbf{Cancer Diagnosis}} \\
\midrule
BRACS-3 & 80.51 (73.63, 87.39) & \underline{90.00} (85.39, 94.61) & 87.34 (81.68, 92.99) & 87.84 (82.90, 92.78)  & \textbf{91.65} (87.18, 96.12) \\
BRACS-7 & 72.68 (66.27, 79.09) & \underline{81.43} (76.18, 86.68) & 77.29 (70.27, 84.31) & 80.85 (75.40, 86.30) & \textbf{85.04} (80.77, 89.31) \\
NSCLC & 90.84 (87.53, 94.15) & 92.89 (89.09, 96.69) & 93.21 (90.43, 96.00) & \underline{94.05} (91.56, 96.54) &  \textbf{95.01} (92.76, 97.26) \\
Camelyon & 86.69 (80.32, 93.06) & 89.86 (83.47, 96.25) & \underline{93.98} (90.33, 97.63) & 90.15 (84.63, 95.67) &  \textbf{95.17} (91.72, 98.62) \\
\midrule
\multicolumn{6}{c}{\textbf{Mutation Prediction}} \\
\midrule
BRCA-TP53 & 60.83 (50.52, 71.14) & 76.23 (67.21, 85.25) & 66.89 (56.89, 76.89) & \underline{77.95} (69.19, 86.71) &  \textbf{80.15} (72.05, 88.25) \\
BRCA-TTN & \underline{42.74} (29.76, 55.72) & 39.75 (23.26, 56.24) & 40.03 (24.98, 55.08) & 42.35 (26.77, 57.93) &  \textbf{52.14} (35.18, 69.10) \\
LUAD-TP53 & 45.61 (33.46, 57.76) & \underline{66.42} (55.35, 77.49) & 60.96 (48.67, 73.25) & 58.55 (46.69, 70.41) & \textbf{68.02} (56.85, 79.19) \\
\midrule
\multicolumn{6}{c}{\textbf{Survival Prediction}} \\
\midrule
BRCA & 59.16 (47.95, 70.37) & \underline{62.50} (51.21, 73.79) & 57.22 (46.00, 68.44) & 62.49 (52.40, 72.58) &  \textbf{64.59} (54.32, 74.86) \\
LUSC & \underline{58.40} (50.07, 66.73) & 55.98 (46.90, 65.06) & 55.50 (47.13, 63.87) & 56.86 (47.40, 66.32) &  \textbf{59.66} (50.31, 69.01) \\
\midrule
\bottomrule
\end{tabular}
}
\label{tab:wsi_cls_scores}
\end{table*}
\begin{table*}[h]
\centering
\caption{Performance of medical vision-language models on zero-shot medical concept annotation tasks (AUC \%). ``w/o Local Info.'' denotes that the local information of image regions is not used in zero-shot concept annnotation. \textbf{Bold} indicates the best result and \underline{underline} indicates the second best. 95\% CI is included in parentheses.
} 
\resizebox{\textwidth}{!}{
\begin{tabular}{l|c|c|c|c|c|c}
\toprule
Model & \makecell{Derm7pt \\ (Dermatology)} & \makecell{SkinCon\\ (Dermatology)}  & \makecell{WBCAtt\\ (Hematology)} & \makecell{BrEaST\\ (Ultrasound)} & \makecell{LUNA16\\ (CT)} & Average\\

\midrule
CLIP
& 53.59 (51.66, 55.32) &	63.88 (62.41, 65.47)&	56.27 (55.61, 56.96)	&45.30 (41.47, 48.97)	&51.64 (49.89, 53.30)	&54.14 \\
SigLIP-400M
& 57.80 (55.70, 59.81)	&67.75 (66.15, 69.62)	&\underline{56.77} (56.14, 57.39)&	51.98 (48.17, 55.68)	&52.54 (50.80, 54.03)	&57.37 \\
PMC-CLIP
& \underline{66.91} (65.09, 68.81)	&61.19 (59.35, 63.06)	&54.93 (54.13, 55.76)	&\underline{56.73} (52.88, 60.42)&	54.84 (53.10, 56.54)	&58.92 \\
BiomedCLIP
&65.28 (63.17, 67.33)&	\underline{68.88} (67.20, 70.57)	&52.44 (51.71, 53.20)	&54.25 (50.28, 58.08)&	\underline{54.99} (53.47, 56.52)	&\underline{59.17}\\
\midrule
\ours{}  
&\textbf{68.56} (66.53, 70.45)	&\textbf{72.20} (70.70, 73.76)	& \textbf{60.59} (59.91, 61.34)&	\textbf{66.24} (61.81, 70.07)&	\textbf{59.04} (57.43, 60.70) &\textbf{65.33}\\
\bottomrule
\end{tabular}
}
\label{tab:zs_concept_anno}
\end{table*}

\begin{table*}[h]
\centering
\caption{Performance of inherently interpretable models built upon medical vision-language models on disease diagnosis tasks (AUC \%). \textbf{Bold} indicates the best result and \underline{underline} indicates the second best. 95\% CI is included in parentheses.
} 
\resizebox{\textwidth}{!}{
\begin{tabular}{l|c|c|c|c|c}
\toprule
Model & SkinCon  & WBCAtt & BrEaST & LUNA16 & Average\\

\midrule
CLIP
&	72.94 (67.36, 78.06)&	95.92 (95.44, 96.34)	&66.22 (50.74, 80.75)	&54.22 (47.26, 60.77)	&72.33 \\
SigLIP-400M
&76.31 (71.17, 81.35)	&\underline{99.15} (98.96, 99.33)&	\underline{74.36} (61.29, 86.89)	&54.24 (47.80, 60.99)	&76.02 \\
PMC-CLIP
&68.33 (62.69, 73.45)	&96.67 (96.29, 97.01)	&71.46 (56.89, 86.79)&	54.00 (47.77, 60.39)	&72.62 \\
BiomedCLIP
&	\underline{77.03} (71.62, 81.66)	&97.97 (97.63, 98.31)	&68.08 (51.77, 81.28)&	\underline{63.94} (57.81, 70.16)	&\underline{76.76}\\
\ours  
&\textbf{80.20}  (75.37, 84.79)	& \textbf{99.49}  (99.33, 99.62)&	\textbf{83.23}  (70.90, 92.95)&	\textbf{68.63} (62.71, 74.42) &\textbf{82.89}\\
\bottomrule
\end{tabular}
}
\label{tab:cbm}
\end{table*}
\begin{table*}[h]
\centering
\caption{{Ablation study on the effectiveness of the RC-Align loss and the local information (``Local Info.'') using the AUC (\%) score. \textbf{Bold} indicates the best results and \underline{underline} indicates the second best. The 95\% CI is included in parentheses.}}
\resizebox{\textwidth}{!}{%
\begin{tabular}{l|c|c|c|c}
\toprule
Model & SIIM-ACR & Covid-CXR2 & VinDr-Mammo & Brain Tumor CT \\ 
\midrule
ConceptCLIP w/o RC-Align & 80.86 (77.87, 83.83) & \underline{80.07} (78.16, 81.77) & 47.72 (43.73, 51.18) & 84.90 (82.42, 87.17) \\ 
ConceptCLIP w/o Local Info. & \underline{81.24} (78.54, 84.04) & 79.67 (77.98, 81.51) & \underline{50.75} (47.41, 54.19) & \underline{90.43} (88.35, 92.28) \\
\midrule
ConceptCLIP & \textbf{83.05} (80.40, 85.63) & \textbf{81.77} (80.19, 83.35) & \textbf{51.78} (48.25, 55.05) & \textbf{92.60} (90.76, 94.20) \\ 
\bottomrule
\end{tabular}
}
\label{tab:zero_shot_cls_ablation_scores}
\end{table*}
\begin{table*}[ht]
\centering
\caption{Medical image analysis dataset splits (training/validation/testing) and evaluation metrics.}
\resizebox{\textwidth}{!}{
\begin{tabular}{|l|c|c|p{6.5cm}|}
\toprule
Task & Dataset & Train/Valid/Test Split & Metrics \\
\midrule
\multirow{12}{*}{Binary Classification} & SIIM-ACR~\cite{siim-acr-pneumothorax-segmentation} & \num{2587} / \num{862} / \num{863} & \multirow{12}{*}{AUC} \\
& Covid-CXR2~\cite{Wang2020} & \num{12648} / \num{4216} / \num{4216} &  \\
& NLM-TB~\cite{jaeger2014two} & \num{480} / \num{160} / \num{160} &  \\
& LC25000 (Colon)~\cite{borkowski2019lung} & \num{6000} / \num{2000} / \num{2000} &  \\
& UBIBC~\cite{ubibc2023} & \num{5514} / \num{1838} / \num{806} &  \\
& PCam200~\cite{kawai2023large} & \num{21405} / \num{7134} / \num{17932} &  \\
& RSNA~\cite{shih2019augmenting} & \num{5777} / \num{1925} / \num{1071} &  \\
\bottomrule
\end{tabular}}
\end{table*}
\fontsize{8}{11}\selectfont
\begin{longtable}{|p{.15\textwidth}|p{.7\textwidth}|p{.05\textwidth}|}
\caption{Medical concept annotation datasets.}\label{tab:datasets_concept} \\
\toprule
Dataset & Description and setting & Link \\
\midrule
\endfirsthead
\multicolumn{3}{c}{{\bfseries \tablename\ \thetable{} -- continued from previous page}} \\
\toprule
Dataset & Description and setting & Link\\
\midrule
\endhead
\midrule
\multicolumn{3}{r}{{Continued on next page}} \\
\bottomrule
\endfoot
\bottomrule
\endlastfoot

Derm7pt~\cite{derm7pt}  & A dermoscopic image dataset containing 1,011 images
with clinical concepts for melanoma skin lesions in dermatology. The test set is adopted for evaluation. Only the dermoscopic images are considered in this paper for concept annotations. The seven clinical concepts include ``Pigment network'', ``Dots and globules'', ``Pigmentation'', ``Streaks'', ``Regression structures'', ``Blue-whitish veil'', and ``Vascular structures''. & \href{https://derm.cs.sfu.ca}{\checkmark}\\
\midrule

SkinCon~\cite{daneshjou2022skincon} & A skin disease dataset densely annotated by experts for fine-grained model debugging and analysis. We choose 22 concepts that
have at least 50 images representing the concept from the Fitzpatrick 17k (F17k) \cite{f17k} subset. 350 images are sampled for evaluation. The clinical concepts include ``Papule'',	``Plaque'',	``Pustule'',	``Bulla'',	``Patch'',	``Nodule'',	``Ulcer'',	``Crust'',	``Erosion'',	``Atrophy'',	``Exudate'',	``Telangiectasia'',	``Scale'',	``Scar'',	``Friable'',	``Dome-shaped'',	``Brown (Hyperpigmentation)'',	``White (Hypopigmentation)'',	``Purple'',	``Yellow'', ``Black'', and ``Erythema''.  & \href{https://skincon-dataset.github.io/}{\checkmark}\\
\midrule

WBCAtt~\cite{tsutsui2024wbcatt} & This dataset is a densely annotated dataset for white blood cell (WBC) recognition, containing 11 morphological attributes for 10,298 cell images. The test set which contains 3,099 images is used for evaluation.  There are 11 concepts, including ``Cell size'', ``Cell shape'', ``Nucleus shape'', ``Nuclear cytoplasmic ratio'',``Chromatin density'', ``Cytoplasm vacuole'', ``Cytoplasm texture'', ``Cytoplasm colour'',	``Granule type'', ``Granule colour'', and ``Granularity''. & \href{https://data.mendeley.com/datasets/snkd93bnjr/1}{\checkmark} \\
\end{longtable}
\begin{table*}[h]
\centering
\caption{State-of-the-art vision-language pre-training models.}
\label{tab:models}
\resizebox{\textwidth}{!}{\begin{tabular}{|p{.30\textwidth}|p{.90\textwidth}|p{.05\textwidth}|}
\toprule
Model Name & Description & Link \\ \midrule
\multicolumn{3}{|c|}{\textbf{Models in the General Domain}} \\ \midrule
CLIP~\cite{radford2021learning} &  \textbf{CLIP}~\cite{radford2021learning} is a vision-language model pre-trained on large-scale image-text pairs via contrastive learning in the general domain, which showcases strong ability in zero-shot classification. We take the version of CLIP-ViT-B-16 pre-trained on LAION-2B~\cite{schuhmann2022laion} under the framework of OpenCLIP~\cite{ilharco_gabriel_2021_5143773}. & \href{https://github.com/openai/CLIP}{\checkmark} \\ \midrule
SigLIP-400M~\cite{zhai2023sigmoid} &  \textbf{SigLIP}~\cite{zhai2023sigmoid} is a variant of the CLIP model, which is pre-trained via contrastive learning using the sigmoid loss. In our experiments, we use the version of ViT-SO400M-14-SigLIP (SigLIP-400M). & \href{https://huggingface.co/docs/transformers/model_doc/siglip}{\checkmark} \\ \midrule
\multicolumn{3}{|c|}{\textbf{Models in the Medical Domain}} \\ \midrule
PMC-CLIP~\cite{lin2023pmc} &  \textbf{PMC-CLIP}~\cite{lin2023pmc} is a vision-language model tailored for medicine. The authors collected PubMed open access articles and conducted a series of data processing operations, producing a dataset with 1.6 million medical image-text pairs. The dataset is used to pre-train a CLIP model, thus resulting in the PMC-CLIP model. We use the origin checkpoint that the authors provided. & \href{https://github.com/WeixiongLin/PMC-CLIP}{\checkmark} \\ \midrule
BiomedCLIP~\cite{zhang2023biomedclip} &  \textbf{BiomedCLIP}~\cite{zhang2023biomedclip} is a medical vision-language model pre-trained on PMC-15M, a dataset with 15 million medical image-text pairs sourced from PMC-OA. We take the released checkpoint for evaluation. & \href{https://huggingface.co/microsoft/BiomedCLIP-PubMedBERT_256-vit_base_patch16_224}{\checkmark} \\ \midrule
\bottomrule
\end{tabular}}
\end{table*}

\end{appendices}

\end{document}